\documentclass{sig-alternate-05-2015}

\usepackage{amssymb}

\usepackage{enumitem}

\usepackage[usenames,dvipsnames]{color}

\usepackage{graphicx}
\usepackage{epstopdf}
\usepackage{amsmath}
\usepackage{bbm}
\usepackage[footnotesize]{caption}
\usepackage{subcaption}
\usepackage{cite}
\usepackage{algorithm}
\usepackage{algorithmicx}
\usepackage[noend]{algpseudocode}
\usepackage{balance} 

\usepackage{color,soul}

\usepackage{tikz}
\usetikzlibrary{arrows}
\usepackage{booktabs}
\usepackage{xifthen}

\newcommand{\xxnote}[3]{}
\ifx\hidenotes\undefined
  \usepackage{color}
  \renewcommand{\xxnote}[3]{\color{#2}{#1: #3}}
\fi

\definecolor{bittersweet}{rgb}{1.0, 0.44, 0.37}
\definecolor{bleudefrance}{rgb}{0.19, 0.55, 0.91}
\definecolor{magenta}{rgb}{1.0, 0.0, 1.0}

\begin{document}

\CopyrightYear{2017} 
\setcopyright{acmcopyright}
\conferenceinfo{HRI '17,}{March 06-09, 2017, Vienna, Austria}
\isbn{978-1-4503-4336-7/17/03}\acmPrice{\$15.00}
\doi{http://dx.doi.org/10.1145/2909824.3020253}

\title{Human-Robot Mutual Adaptation in Shared Autonomy 
}

\numberofauthors{4}
  \author{
      \alignauthor Stefanos Nikolaidis\\ 
          \affaddr{Carnegie Mellon University} \\            
      \email{snikolai@cmu.edu}
      \alignauthor Yu Xiang Zhu\\    
          \affaddr{Carnegie Mellon University} \\            
      \email{yuxiangz@cmu.edu}
  \and
      \alignauthor 
    David Hsu\\                       
    \affaddr{National University of Singapore} \\            
    \email{dyhsu@comp.nus.edu.sg}  
      \alignauthor 
    Siddhartha Srinivasa\\                       
    \affaddr{Carnegie Mellon University} \\            
    \email{siddh@cmu.edu}
}

\maketitle
\begin{abstract}
  Shared autonomy integrates user input with robot autonomy in order to   control a robot and help the user to complete a task. Our work aims to   improve the performance of such a human-robot team: the robot tries to guide
  the human towards an effective strategy, sometimes against the human's own preference, while still retaining his trust. We achieve this through a principled human-robot mutual adaptation formalism. We integrate a bounded-memory adaptation model of the human into a partially observable stochastic decision model, which enables the robot to adapt to an adaptable human. When the human is adaptable, the robot guides the human towards a good strategy, maybe unknown to the human in advance. When the human is stubborn and not adaptable, the robot complies with the human's preference in order to retain their trust.  In the shared autonomy setting, unlike many other common human-robot collaboration settings, only the robot actions can change the physical state of the world, and the human and robot goals are not fully observable.  We address these challenges and show in a human subject experiment that the proposed mutual adaptation formalism improves human-robot team performance, while retaining a high level of user trust in the robot, compared to the common approach of having the robot strictly following participants' preference.


\end{abstract}

\section{Introduction}

\begin{figure}[t!]
\centering
  \includegraphics[width=1.0\linewidth]{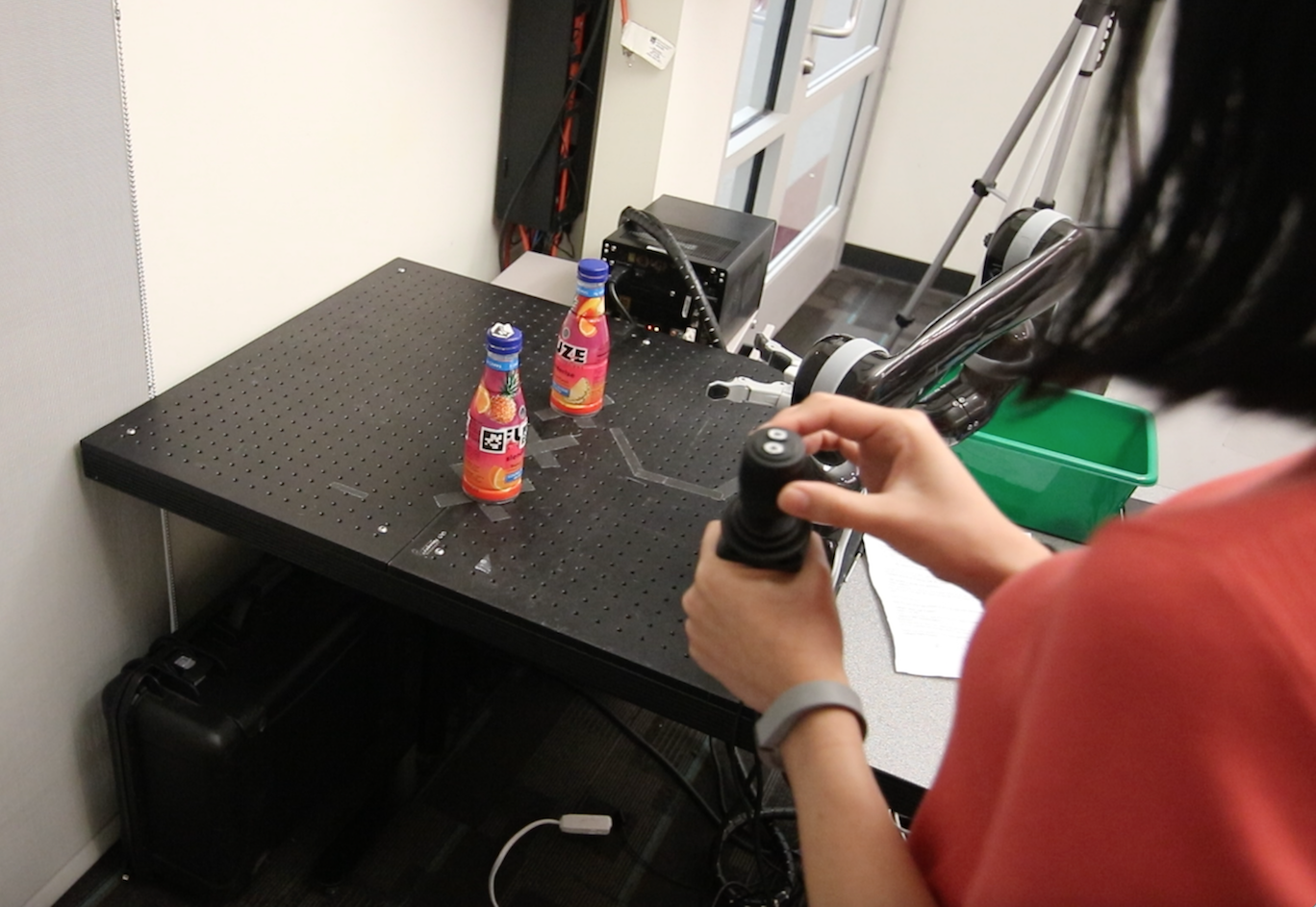}
  \includegraphics[width=1.0\linewidth]{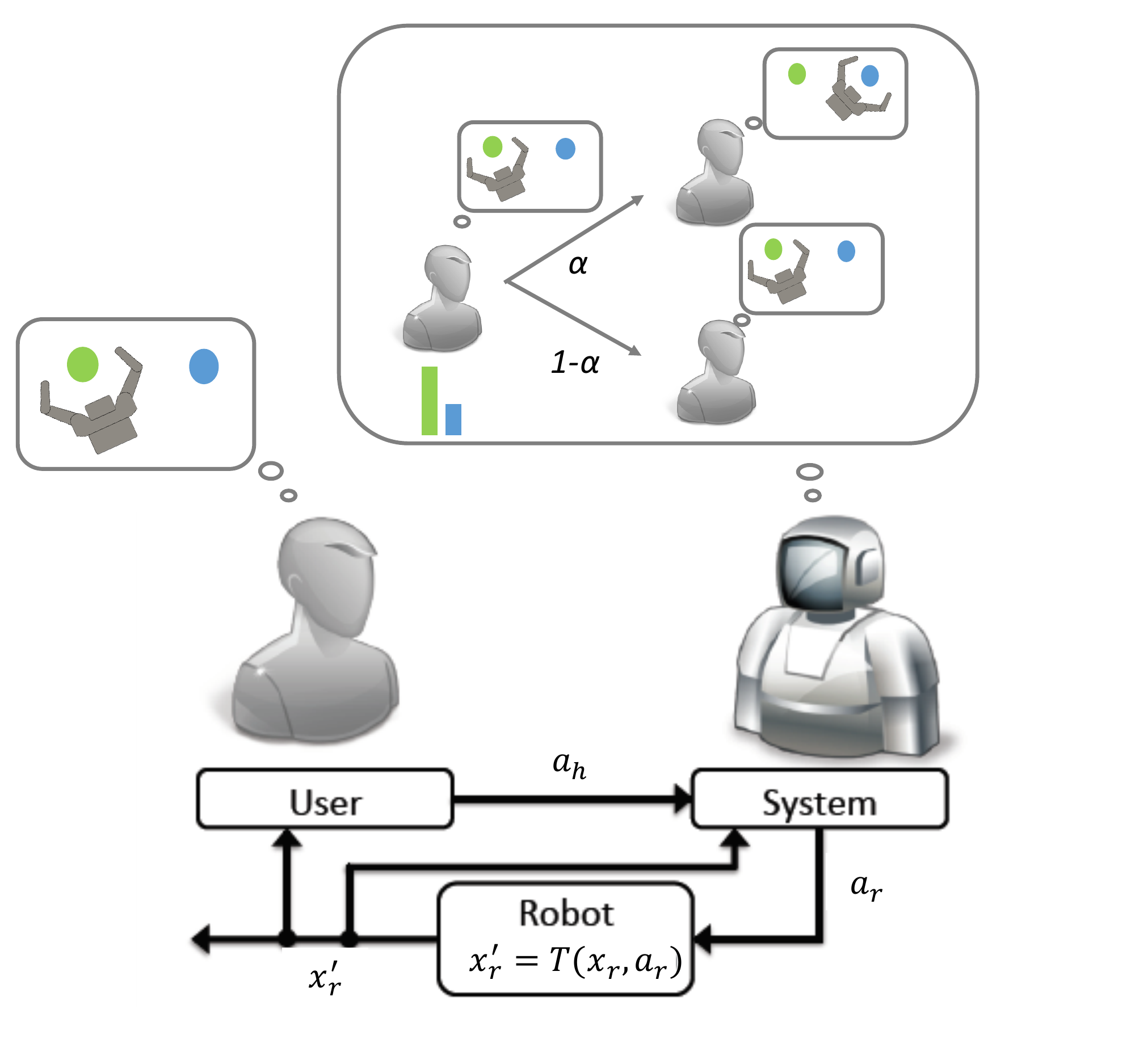}
 \caption{Table clearing task in a shared autonomy setting. The user operates the robot using a joystick interface and moves the robot towards the left bottle, which is a suboptimal goal. The robot plans its actions based on its estimate of the current human goal and the probability $\alpha$ of the human switching towards a new goal indicated by the robot.}
 \label{fig:screenshot}
\end{figure}

\begin{figure*}[t!]
\centering

\setlength\tabcolsep{1.5pt}
\centering
\begin{tabular}{cccc}
\begin{subfigure}[l]{0.23\linewidth}
\centering
\includegraphics[height=3.0cm]{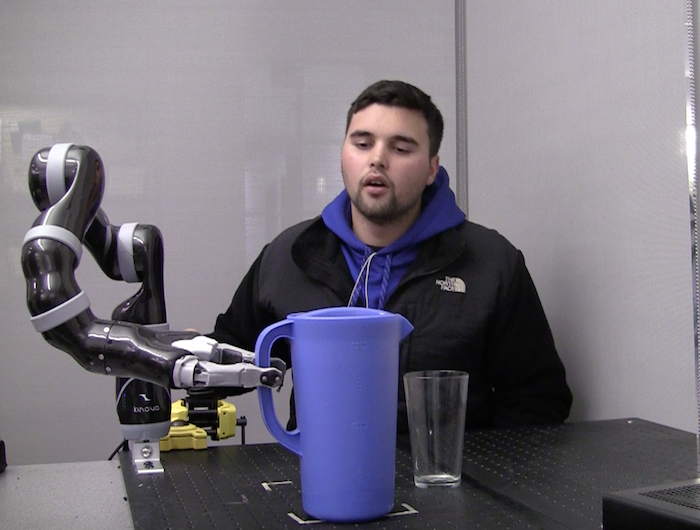} \\
\label{fig:path-left}
\end{subfigure}
&
\begin{subfigure}[l]{0.23\linewidth}
\centering
\includegraphics[height=3.0cm]{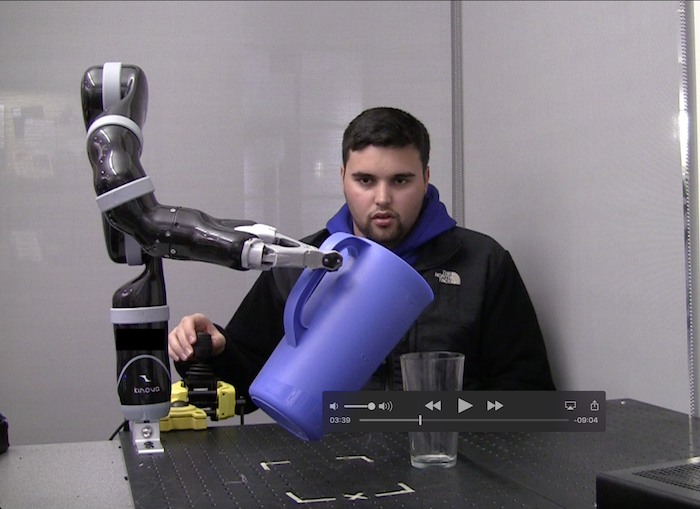} \\
\label{fig:path-right}
\end{subfigure}
&
\begin{subfigure}[l]{0.23\linewidth}
\centering
\includegraphics[height=3.0cm]{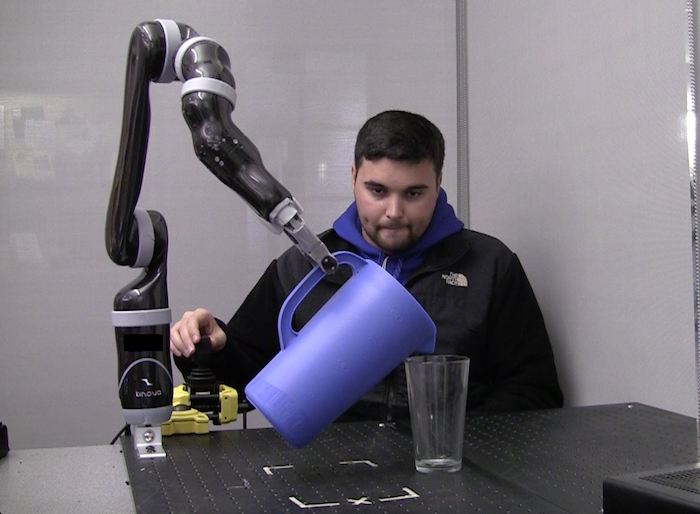} \\
\label{fig:straight-path}
\end{subfigure}
&
\begin{subfigure}[l]{0.23\linewidth}
\centering
\includegraphics[height=3.0cm]{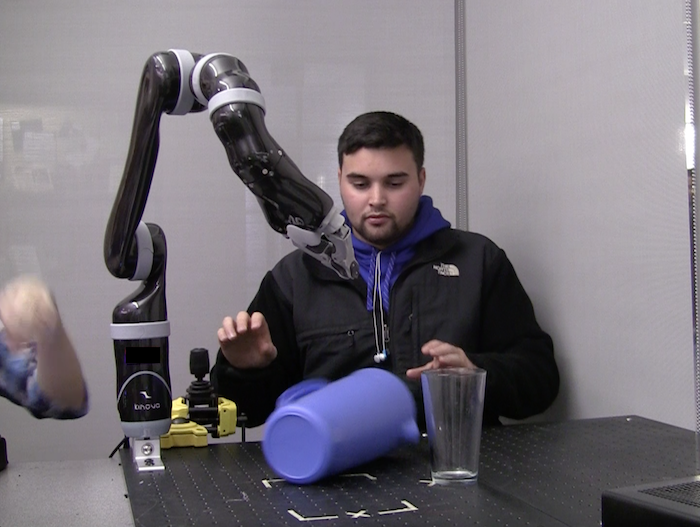} \\
\label{fig:straight-path}
\end{subfigure}
\end{tabular}
\caption{The user guides the robot towards an unstable grasp, resulting in task failure.}
\label{fig:grasp-strategy}
\end{figure*}

Assistive robot arms show great promise in increasing the independence of
people with upper extremity disabilities~\cite{hillman2002weston,
  prior1990electric,sijs2007combined}. However, when a user teleoperates
directly a robotic arm via an interface such as a joystick, the limitation of
the interface, combined with the increased capability and complexity of robot
arms, often makes it difficult or tedious to accomplish complex tasks.

Shared autonomy alleviates this issue by combining direct teleoperation with
autonomous assistance~\cite{kofman2005teleoperation, dragan2013policy,
  yu2005telemanipulation, trautman2015assistive,gopinath2017human}. In recent
work by Javdani et al., the robot estimates a distribution of user goals based on the history of user inputs, and assists the user for that distribution~\cite{Javdani-RSS-15}. The user is assumed to be always right about their goal choice. Therefore, if the assistance strategy knows the user's goal, it will select actions to minimize the cost-to-go to that goal. 
This assumption is often not true, however. For instance, a user may choose an unstable grasp when picking up an object~(Fig.~\ref{fig:grasp-strategy}), or they may arrange items in the wrong order by stacking a heavy item on top of a fragile one.  Fig.~\ref{fig:screenshot} shows a shared autonomy scenario, where the user teleoperates the robot towards the left bottle. We assume that the robot \textit{knows the optimal goal for the task}: picking up the right bottle is a better choice, for instance because the left bottle is too heavy, or because the robot has less uncertainty about the right bottle's location. Intuitively, if the human insists on the left bottle, the robot should comply; failing to do so can have a negative effect on the user's trust in the robot, which may lead to disuse of the system~\cite{hancock2011meta,salem2015would,lee2013computationally}. If the human is willing to adapt by aligning its actions with the robot, which has been observed in adaptation between humans and artifacts~\cite{xu2009woz,komatsu2005experiments}, the robot should insist towards the optimal goal. The human-robot team then exhibits \textit{a mutually adaptive behavior, where the robot adapts its own actions by reasoning over the adaptability of the human teammate}. 

Nikolaidis et al.~\cite{nikolaidis2016formalizing} proposed a mutual
adaptation formalism in collaborative tasks, e.g., when a human and a robot work together to carry a table out of the room. The robot builds a Bounded-memory Adaptation Model (BAM) of the human teammate, and it integrates the model into a partially observable stochastic process, which enables robot adaptation to the human: If the user is adaptable, the robot will disagree with them, expecting them to switch towards the optimal goal. Otherwise, the robot will align its actions with the user policy, thus retaining their trust. 

A characteristic of many collaborative settings is that human and robot both affect the world state, and that disagreement between the human and the robot impedes task completion. For instance, in the table-carrying example, if human and robot attempt to move the table in opposing directions with equal force, the table will not move and the task will not progress. Therefore, a robot that optimizes the task completion time will account for the human adaptability implicitly in its optimization process: if the human is non-adaptable, the only way for the robot to complete the task is to follow the human goal. This is not the case in a shared-autonomy setting, since the human actions do not affect the state of the task. Therefore, a robot that solely maximizes task performance in that setting will always move towards the optimal goal, ignoring human inputs altogether. 




In this work, \textit{we propose a generalized human-robot mutual adaptation formalism}, and we formulate mutual adaptation in the collaboration and shared-autonomy settings as instances of this formalism. 

We identify that in the shared-autonomy setting (1) tasks may typically exhibit less structure than in the collaboration domain, which limits the observability of the user's intent, and (2) only robot actions directly affect task progress. We address the first challenge by including the operator goal as an additional latent variable in a mixed-observability Markov decision process (MOMDP)~\cite{ong2010planning}. This allows the robot to maintain a probability distribution over the user goals based on the history of operator inputs. We also take into account the uncertainty that the human has on the robot goal by modeling the human as having a probability distribution over the robot goals (Sec.~\ref{sec:partially-observable-modes}). We address the second challenge by proposing an explicit penalty for disagreement in the reward function that the robot is maximizing  (Sec.~\ref{sec:optimal-disagreement}). This allows the robot to infer simultaneously the human goal and the human adaptability, reason over how likely the human is to switch their goal based on the robot actions, and guide the human towards the optimal goal while retaining their trust.

 We conducted a human subject experiment ($n=51$) with an assistive robotic arm on a table-clearing task.  Results show that the proposed formalism significantly improved human-robot team performance, compared to the robot following participants' preference, while retaining a high level of human trust in the robot.


\section{Problem Setting} \label{sec:problem}
A human-robot team can be treated as a multi-agent system, with world state $x_{world} \in X_{world}$, robot action $a_r \in A_r$, and human action $a_h \in A_h$. The system evolves according to a stochastic state transition function $T\colon X_{world} \times A_r \times A_h \rightarrow \Pi(X_{world})$. Additionally, we model the user as having a goal, among a discrete set of goals $g \in G$. We assume access to a stochastic joint policy for each goal, which we call \textit{modal policy}, or \textit{mode} $m \in M$. We call $m_h$ the modal policy that the human is following at a given time-step, and $m_r$ the robot mode, which is the [perceived by the human] robot policy towards a goal. The human mode, $m_h \in M$ is not fully observable. Instead, the robot has uncertainty over the user's policy, that can modeled as a Partially Observable Markov Decision Process (POMDP). Observations in the POMDP correspond to human actions $a_h \in A_h$. Given a sequence of human inputs, we infer a distribution over  user modal policies using an observation function $O(a_h | x_{world}, m_h)$.

Contrary to previous work in modeling human intention~\cite{bandyopadhyay2013intention} and in shared autonomy~\cite{Javdani-RSS-15}, the user goal is not static. Instead, we define a transition function $T_{m_h}\colon M \times H_t \times X_{world} \times A_r \rightarrow \Pi(M)$, where $h_t$ is the history of states, robot and human actions $\bigl(x^0_{world}, a^0_r,a^0_h, \ldots , x^{t-1}_{world}, a^{t-1}_r, a^{t-1}_h \bigr)$. The function models how the human mode may change over time. At each time step, the human-robot team receives a real-valued reward that in the general case also depends on the human mode $m_h$ and history $h_t$: $R(m_h, h_t, x_{world}, a_r, a_h)$. The reward captures both the relative cost of each goal $g \in G$, as well as the cost of disagreement between the human and the robot. The robot goal is then  to maximize the expected total reward over time: ${\sum_{t=0}^{\infty} {\gamma^t R(t)}}$, where the discount factor $\gamma\in [\,0,1)$ gives higher weight to immediate rewards than future ones.

 Computing the maximization is hard: Both $T_{m_h}$ and $R$ depend on the whole history of states, robot and human actions $h_t$. We use the Bounded-memory Adaptation Model~\cite{nikolaidis2016formalizing} to simplify the problem.

\subsection{Bounded-Memory Adaptation Model} \label{subsec:BAM}

Nikolaidis et al.~\cite{nikolaidis2016formalizing} proposed the Bounded-memory Adaptation Model (BAM). The model is based on the assumption of ``bounded rationality'' was proposed first by Herbert Simon: people often do not have the time and cognitive capabilities to make perfectly rational decisions~\cite{simon1979rational}. In game theory,  bounded rationality has been modeled by assuming that players have a ``bounded memory'' or ``bounded recall" and base their decisions on recent observations~\cite{powers2005learning, monte2014learning,aumann1989cooperation}.

The BAM model allows us to simplify the problem, by modeling the human as making decisions not on the whole history of interactions, but on the last $k$ interactions with the robot. This allows us to simplify the transition function $T_{m_h}$ and reward function $R$ defined in the previous section, so that they depend on the history of the last $k$ time-steps only. Additionally, BAM provides a parameterization of the transition function $T_{m_h}$, based on the parameter $\alpha \in \mathcal{A}$, which is the human \textit{adaptability}. The adaptability represents one's inclination to adapt to the robot. With the BAM assumptions, we have $T_{m_h}\colon M \times \mathcal{A} \times H_k \times X_{world} \times A_r \rightarrow \Pi(M)$ and $R: M \times H_k \times X_{world} \times A_r \times A_h \rightarrow \mathbb{R}$. We describe the implementation of $T_{m_h}$ in Sec.~\ref{sec:partially-observable-modes} and of $R$ in Sec.~\ref{sec:optimal-disagreement}.


\subsection{Shared Autonomy}
The shared autonomy setting allows us to further simplify the general problem: the world state consists only of the robot configuration $x_r \in X_r$, so that $x_r \equiv x_{world}$. A robot action induces a deterministic change in the robot configuration. The human actions  $a_h \in A_h$ are inputs through a joystick interface and do not affect the world state. Therefore, the transition function of the system is deterministic and can be defined as: $T\colon X_{r} \times A_r  \rightarrow X_{r}$.

\subsection{Collaboration}
We include the collaboration setting formalism for completeness. Contrary to the shared-autonomy setting, both human and robot actions affect the world state, and the transition function can be deterministic or stochastic. In the deterministic case, it is $T: X_{world} \times A_r \times A_h \rightarrow X_{world}$. Additionally, the reward function does not require a penalty for disagreement between the human and robot modes; instead, it can depend only on the relative cost for each goal, so that $R : X_{world} \rightarrow \mathbb{R}$. Finally, if the task exhibits considerable structure, the modes may be directly observed from the human and robot actions. In that case, the robot does not maintain a distribution over modal policies. 

\section{Human and Robot Mode Inference} \label{sec:partially-observable-modes}
When the human provides an input through a joystick interface, the robot makes an inference on the human mode. In the example table-clearing task of Fig.~\ref{fig:screenshot}, if the robot moves to the right, the human will infer that the robot follows a modal policy towards the right bottle. Similarly, if the human moves the joystick to the left, the robot will consider more likely that the human follows a modal policy towards the left bottle. In this section, we formalize the inference that human and robot make on each other's goals.

\subsection{Stochastic Modal Policies} \label{subsec:stochastic}

\begin{figure}[t!]
\centering
 \fbox{\includegraphics[width=0.4\linewidth]{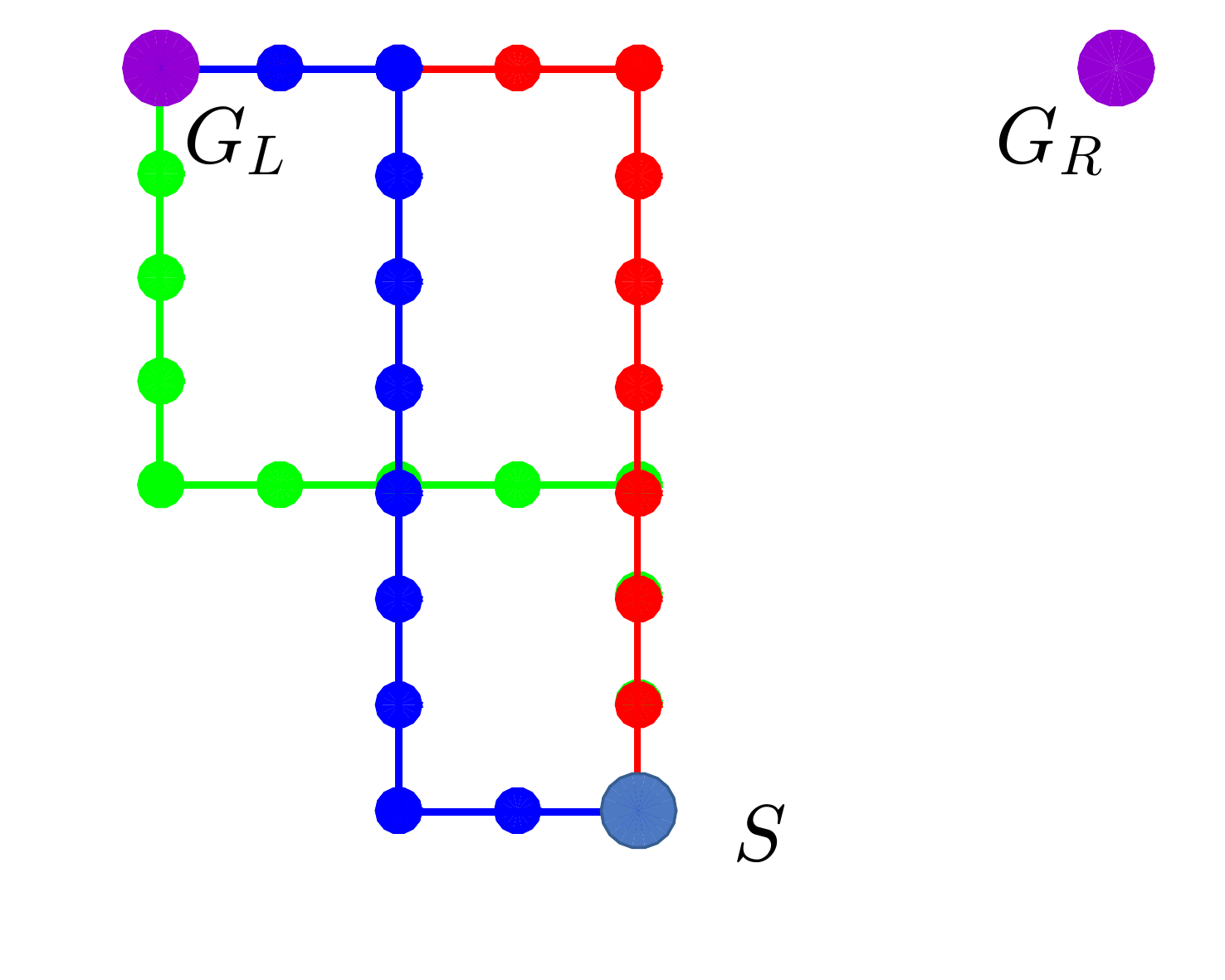}}
   \hspace{0.4cm}\fbox{\includegraphics[width=0.4\columnwidth]{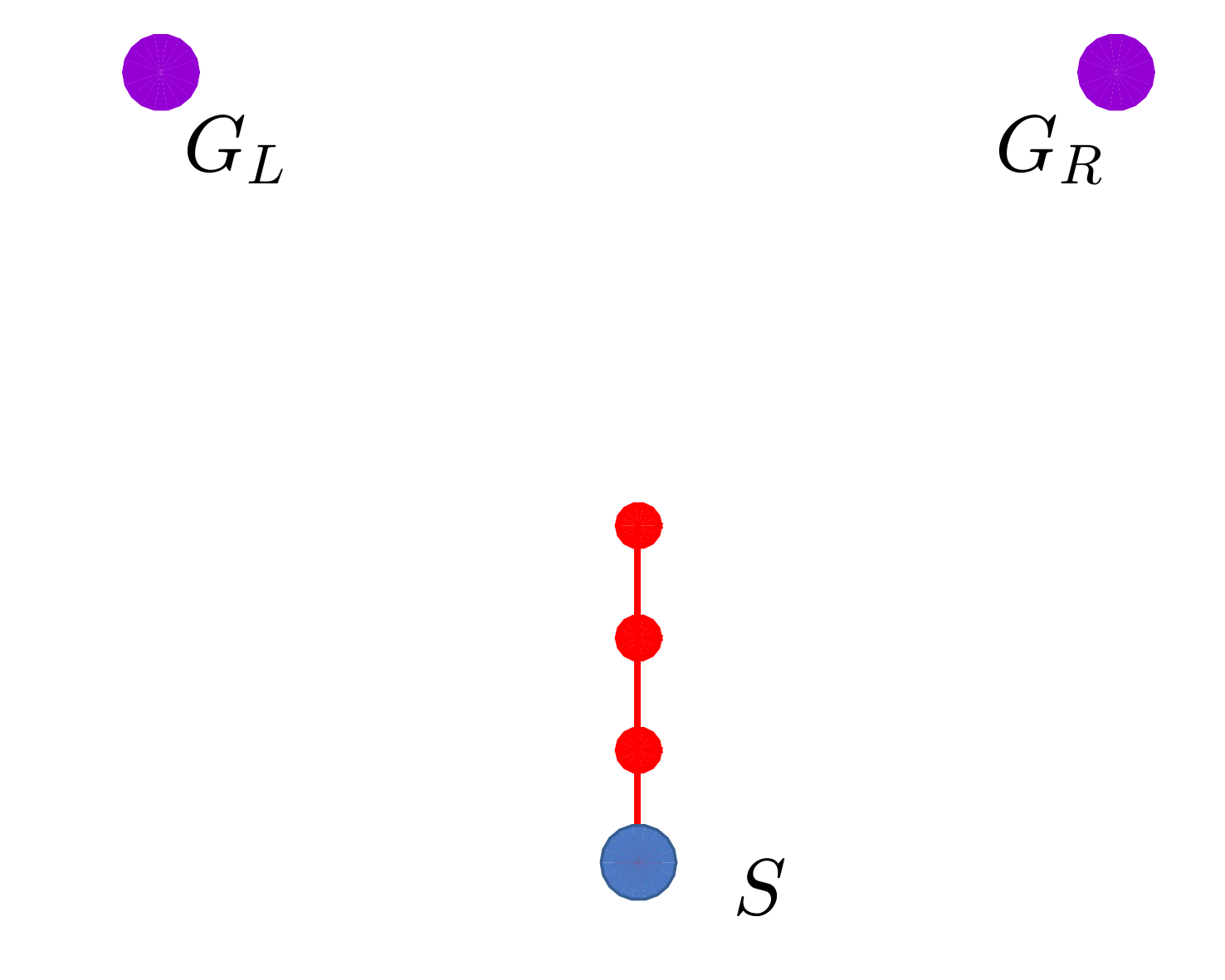}}

 \caption{(left) Paths corresponding to three different modal policies that lead to the same goal $G_L$. We use a stochastic modal policy $m_L$ to compactly represent all feasible paths from $S$ to $G_L$. (right) The robot moving upwards from point $S$ could be moving towards either $G_L$ or $G_R$.}
 \label{fig:MultipleTrajs}
\end{figure}

In the shared autonomy setting, there can be a very large number of modal policies that lead to the same goal. We use as example the table-clearing task of Fig.~\ref{fig:screenshot}. We let $G_L$ represent the left bottle, $G_R$ the right bottle, and $S$ the starting end-effector position of the robot. Fig.~\ref{fig:MultipleTrajs}-left shows paths from three different modal policies that lead to the same goal $G_L$. Accounting for a large set of modes can increase the computational cost, in particular if we assume that the human mode is partially observable (Section~\ref{sec:formalism}). 



Therefore, we define a modal policy as a stochastic joint-policy over human and robot actions, so that $m\colon X_r \times H_{t} \rightarrow \Pi(A_r) \times \Pi(A_h)$. A stochastic modal policy compactly represents a probability distribution over paths and allows us to reason probabilistically about the future actions of an agent that does not move in a perfectly predictable manner. For instance, we can use the principle of maximum entropy to create a probability distribution over all paths from start to the goal~\cite{ziebart2009planning,ziebart2008maximum}. While a stochastic modal policy represents the uncertainty of the observer over paths, we do not require the agent to actually follow a stochastic policy.

\subsection{Full Observability Assumption}
 While $m_r$, $m_h$ can be assumed to be observable for a variety of structured tasks in the collaboration domain~\cite{ nikolaidis2016formalizing}, this is not the case for the shared autonomy setting because of the following factors:
\\
\noindent{\bf Different policies invoke the same action.} Assume two modal policies in Fig.~\ref{fig:MultipleTrajs}, one for the left goal shown in red in Fig.~\ref{fig:MultipleTrajs}-left, and a symmetric policy for the right goal (not shown). An agent moving upwards (Figure~\ref{fig:MultipleTrajs}-right) could be following either of the two with equal probability. In that case, inference of the exact modal policy without any prior information is impossible, and the observer needs to maintain a uniform belief over the two policies. 

\noindent{\bf Human inputs are noisy.}  The user provides its inputs to the system through a joystick interface. These inputs are noisy: the user may ``overshoot'' an intended path and correct their input, or move the joystick in the wrong direction. In the fully observable case, this would result in an incorrect inference of the human mode. Maintaining a belief over modal policies  allows robustness to human mistakes.

This leads us to assume that modal policies are \textit{partially observable.} We model how the human infers the robot mode, as well as how the robot infers the human mode, in the following sections.

\subsection{Robot Mode Inference} \label{subsec:RobotModeInference}

The bounded-memory assumption dictates that the human does not recall the whole history of states and actions, but only a recent history of the last $k$ time-steps. The human attributes the robot actions to a robot mode $m_r$.





\begin{equation}
\begin{split}
P(m_r | h_k, x_r^{t}, a_r^{t}) & = P (m_r | x_r^{t-k+1}, a_r^{t-k+1}, ... , x_r^{t}, a_r^{t}) \\ 
                 &= \eta~P (a_r^{t-k+1}, ... , a_r^{t}| m_r, x_r^{t-k+1}, ... , x_r^{t}) 
\end{split}
\label{e:m_r}
\end{equation}

In this work, we consider modal policies that generate actions based only on the current world state, so that
$M : X_r \rightarrow \Pi(A_h) \times \Pi(A_r)$. 

Therefore Eq.~\ref{e:m_r} can be simplified as follows, where $m_r(x^t_r,a^t_r)$ denotes the probability of the robot taking action $a_r$ at time $t$, if it follows modal policy $m_r$: 
\begin{equation}
\begin{split}
P(m_r | h_k, x_r^{t}, a_r^{t}) &= \eta~m_r(x_r^{t-k+1},a_r^{t-k+1}) ... m_r(x_r^{t},a_r^{t})  \\
\end{split}
\label{e:m_r_simple}
\end{equation}

$P(m_r | h_k,x_r^{t},a_r^{t})$ is the [estimated by the robot] human belief on the robot mode $m_r$. 

\subsection{Human Mode Inference} \label{subsec:human-mode-inference}
To infer the human mode, we need to implement the dynamics model $T_{m_h}$ that describes how the human mode evolves over time, and the observation function $O$ that allows the robot to update its belief on the human mode from the human actions.

In Sec.~\ref{sec:problem} we defined a transition function $T_{m_h}$, that indicates the probability of the human switching from mode $m_h$ to a new mode $m'_h$, given a history $h_k$ and their adaptability $\alpha$. We simplify the notation, so that $x_r \equiv x^t_{r}$, $a_r \equiv a^t_r$ and $x \equiv (h_k, x_r)$:

\begin{equation}
\begin{split}
&T_{m_h}(x, \alpha,m_h, a_r, m'_h)
= P(m'_h | x, \alpha, m_h,a_r) \\ 
& =\sum_{m_r} P(m'_h, m_r |x,\alpha,m_h,a_r) \\
&= \sum_{m_r} P(m'_h | x,\alpha,m_h, a_r,m_r) \times P(m_r | x,\alpha,m_h,a_r)\\
&= \sum_{m_r} P(m'_h | \alpha, m_h, m_r)\times P(m_r | x,a_r) 
\end{split}
\label{e:m_h-transition}
\end{equation}
The first term gives the probability of the human switching to a new mode $m'_h$, if the human mode is $m_h$ and the robot mode is $m_r$. 
Based on the BAM model~\cite{nikolaidis2016formalizing}, the human switches to $m_r$, with probability~$\alpha$ and stays at $m_h$ with probability $1-\alpha$. Nikolaidis et al. \cite{nikolaidis2016formalizing} define $\alpha$ as the human adaptability, which represents their inclination to adapt to the robot. If $\alpha=1$, the human switches to $m_r$ with certainty. If $\alpha=0$, the human insists on their mode $m_h$ and does not adapt. Therefore:

\begin{equation}
P(m'_h|\alpha,m_h,m_r) = \left\{\begin{matrix}
 \alpha & m'_h \equiv m_r\\ 
 1-\alpha & m'_h \equiv m_h\\ 
 0 & \text{otherwise}\\ 
\end{matrix}\right.
\end{equation}


The second term in Eq.~\ref{e:m_h-transition} is computed using Eq.~\ref{e:m_r_simple}, and it is the [estimated by the human] robot mode.  

Eq.~\ref{e:m_h-transition} describes that the probability of the human switching to a new robot mode $m_r$ depends on the human adaptability $\alpha$, as well as on the uncertainty that the human has about the robot following $m_r$. This allows the robot to compute the probability of the human switching to the robot mode, given each robot action.

The observation function $O \colon X_{r} \times M \rightarrow \Pi(A_h)$ defines a probability distribution over human actions $a_h$. This distribution is specified by the stochastic modal policy $m_h \in M$. Given the above, the human mode $m_h$ can be estimated by a Bayes filter, with $b(m_h)$ the robot's previous belief on $m_h$: 
\begin{equation}
\label{e:beliefupdate2a}
\begin{split}
b'(m'_h) = &\eta~O(m'_h, x'_r, a_h)\sum_{m_h \in M}T_{m_h}(x, \alpha, m_h,a_r, m'_h) b(m_h)
\end{split}
\end{equation}

In this section, we assumed that $\alpha$ is known to the robot. In practice, the robot needs to estimate both $m_h$ and $\alpha$. We formulate this in Sec.~\ref{sec:formalism}.

\section{Disagreement between Modes} \label{sec:optimal-disagreement}
In the previous section we formalized the inference that human and robot make on each other's goals. Based on that, the robot can infer the human goal and it can reason over how likely the human is to switch goals given a robot action.

Intuitively, if the human insists on their goal, the robot should follow the human goal, even if it is suboptimal, in order to retain human trust. If the human is willing to change goals, the robot should move towards the optimal goal. We enable this behavior by proposing in the robot's reward function  a penalty for disagreement between human and robot modes. The intuition is that if the human is non-adaptable, they will insist on their own mode throughout the task, therefore the expected accumulated cost of disagreeing with the human will outweigh the reward of the optimal goal. In that case, the robot will follow the human preference. If the human is adaptable, the robot will move towards the optimal goal, since it will expect the human to change modes.


We formulate the reward function that the robot is maximizing, so that there is a penalty for following a mode that is perceived to be different than the human's mode. 


\begin{align}
    R(x, m_h, a_r) = \begin{cases}
        R_{goal}        &: x_r \in G \\
        R_{other}    &: x_r  \notin G
    \end{cases}
\label{e:reward_full}
\end{align}

 If the robot is at a goal state $x_r \in G$, a positive reward associated with that goal is returned, regardless of the human mode $m_h$ and robot mode $m_r$. Otherwise, there is a penalty $C<0$ for disagreement between $m_h$ and  $m_r$, induced in $R_{other}$. The human does not observe $m_r$ directly, but estimates it from the recent history of robot states and actions (Sec.~\ref{subsec:RobotModeInference}). Therefore, $R_{other}$ is computed so that the penalty for disagreement is weighted by the [estimated by the human] probability of the robot actually following $m_r$:

\begin{equation}
\label{e:reward}
\begin{split}
R_{other} &= \sum_{m_r}R_m(m_h,m_r)P(m_r | x,a_r)                
\end{split}
\end{equation}


\begin{align}
   \text{where } R_m(m_h, m_r) = \begin{cases}
        0         &: m_h \equiv m_r \\
        C        &: m_h \neq m_r 
    \end{cases}
\label{e:disagreement_cost}
\end{align}

\begin{figure*}[t!]
\centering

\setlength\tabcolsep{1.5pt}

\begin{subfigure}[b]{1.0\linewidth}
\begin{tabular}{ccccccc}
\begin{subfigure}[l]{.065\linewidth}
\centering
  \small User 1 \\ $\alpha = 0.0$ $m = m_{L}$ \\
  \vspace{6cm}
\end{subfigure}
\begin{subfigure}[b]{.18\linewidth}
\centering
  \begin{tabular}{cc}
    ~~~\small $T = 1$\\
    ~~~~~\includegraphics[width=0.9\linewidth]{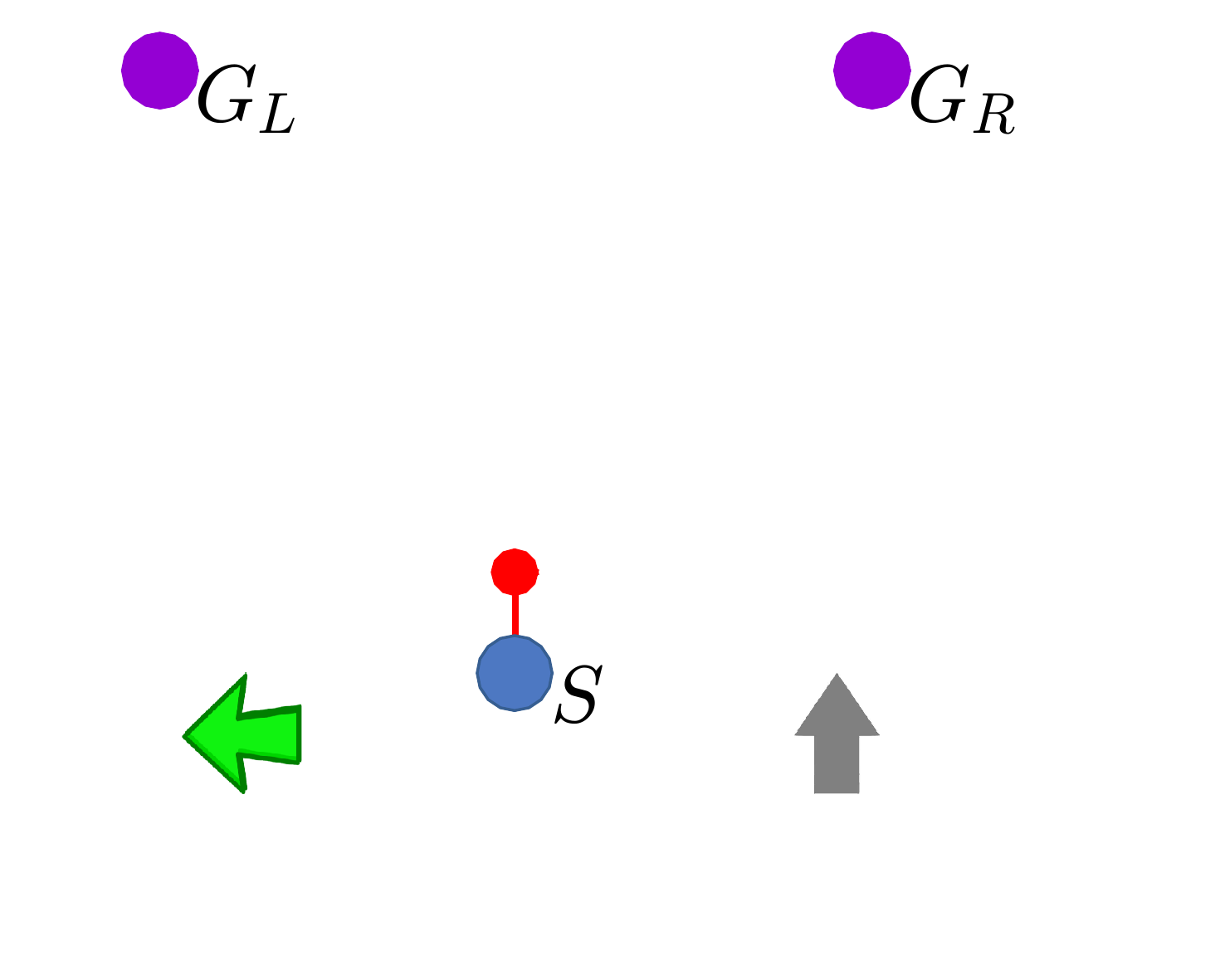}\vspace{-0.3cm}\\
 \includegraphics[width=0.9\linewidth]{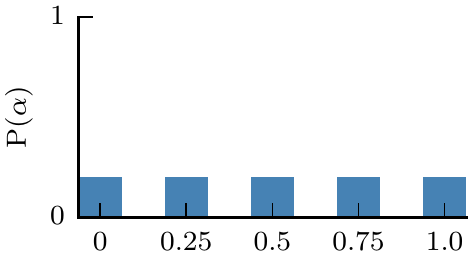}\\
 \includegraphics[width=0.9\linewidth]{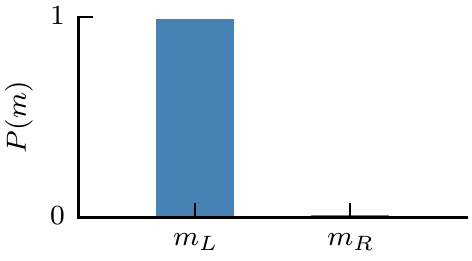}\\
   \end{tabular}
 \label{fig:plotT0}
\end{subfigure}
&
\begin{subfigure}[b]{.18\linewidth}
\centering
  \begin{tabular}{cc}
    ~~~\small $T = 2$\\
   ~~~~~\includegraphics[width=0.9\linewidth]{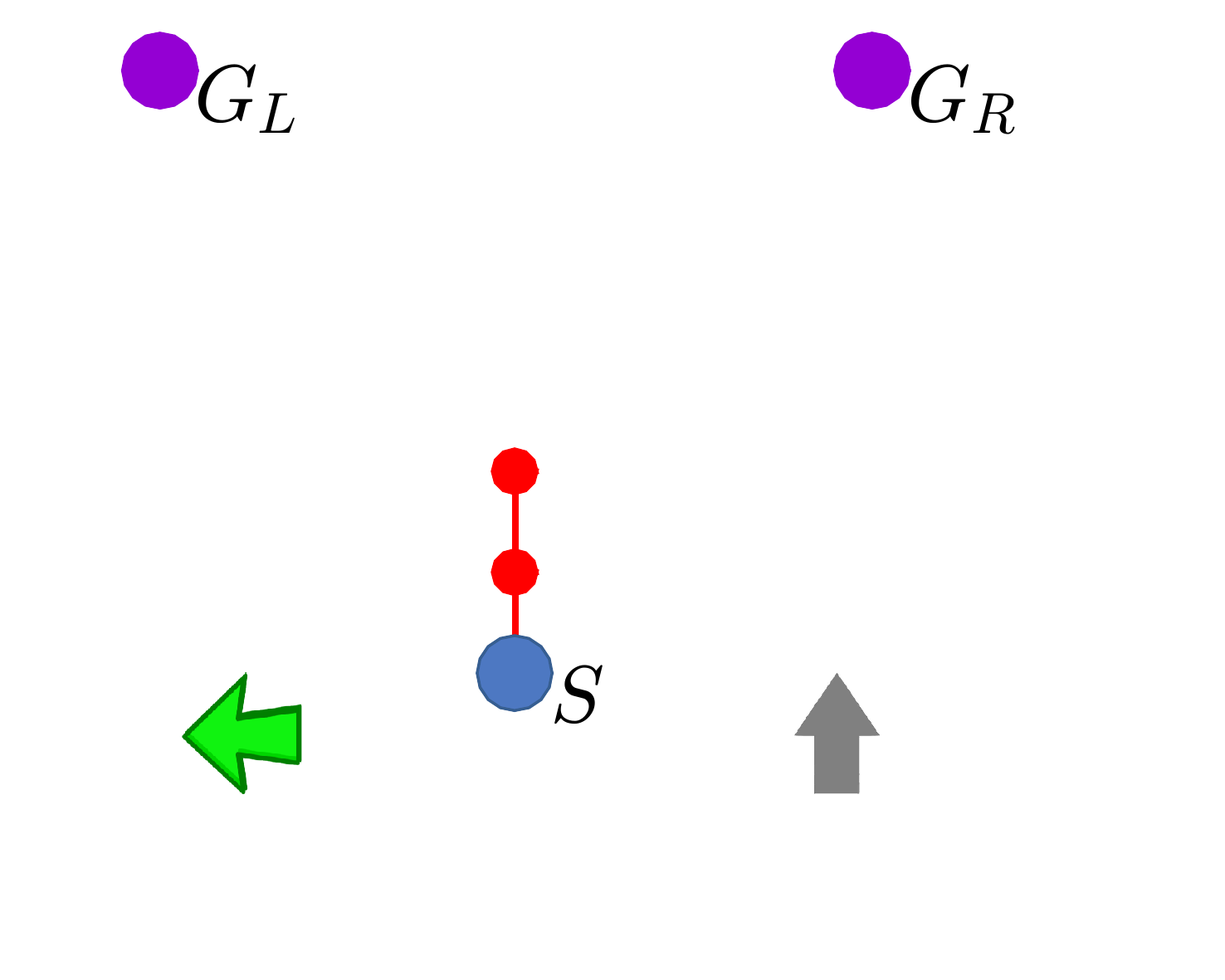}\vspace{-0.3cm}\\
 \includegraphics[width=0.9\linewidth]{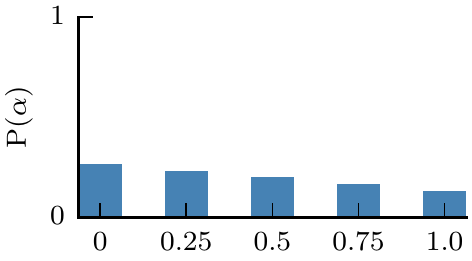}\\
  \includegraphics[width=0.89\linewidth]{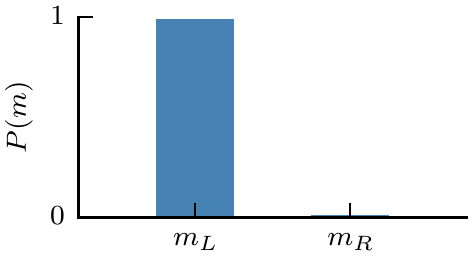}\\
   \end{tabular}
 \label{fig:plotT0}
\end{subfigure}
\begin{subfigure}[b]{.18\linewidth}
\centering
  \begin{tabular}{cc}
    ~~~\small $T = 3$\\
      ~~~~~\includegraphics[width=0.9\linewidth]{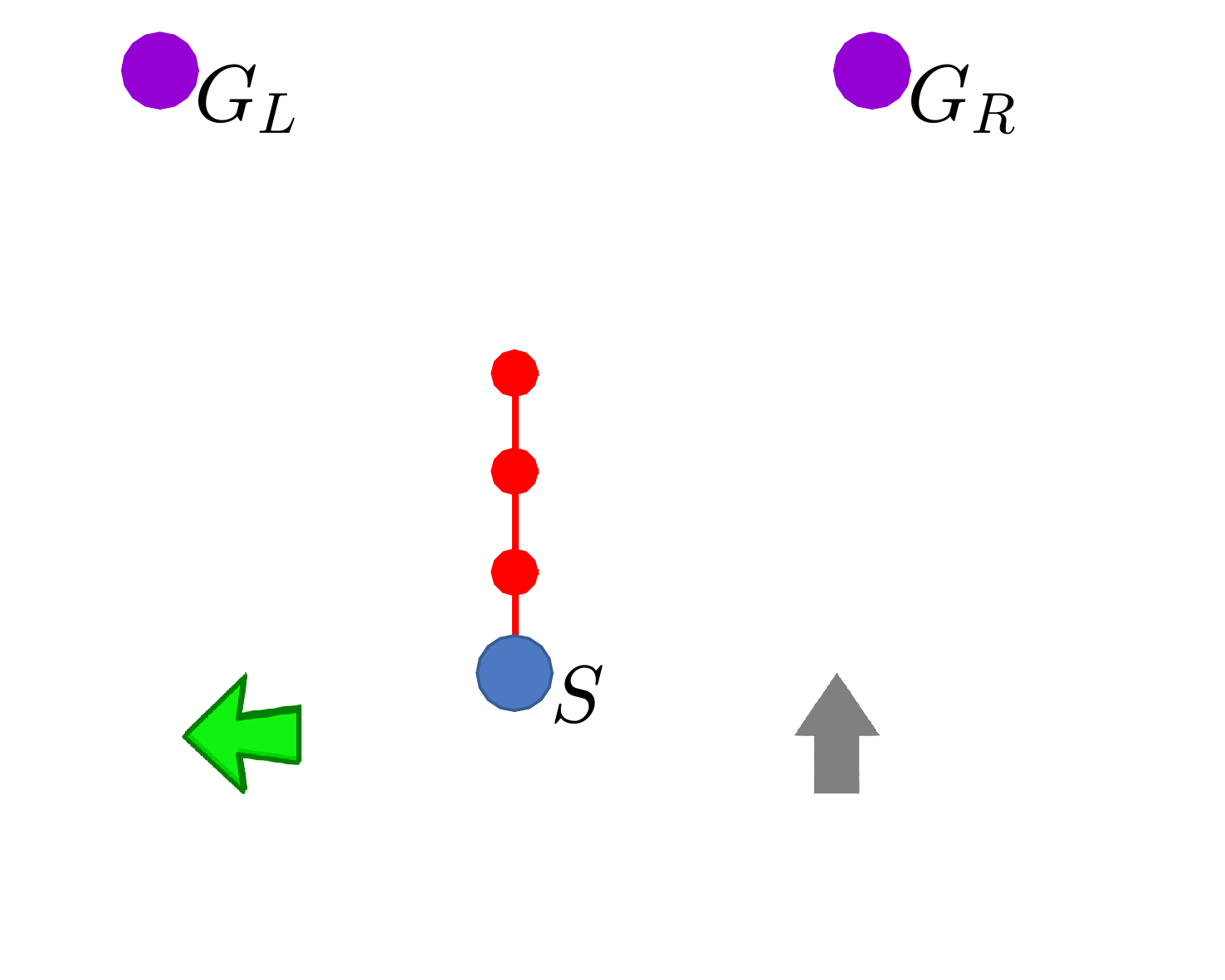}\vspace{-0.3cm}\\
 \includegraphics[width=0.9\linewidth]{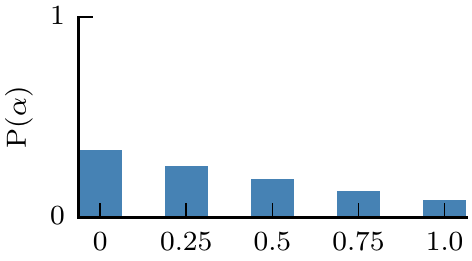}\\
 \includegraphics[width=0.9\linewidth]{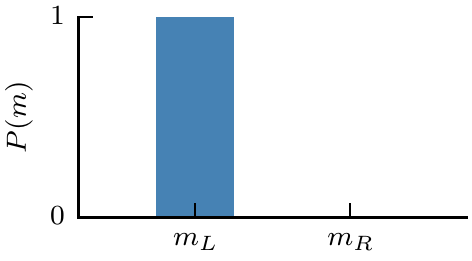}\\
   \end{tabular}
 \label{fig:plotT0}
\end{subfigure}
&
\begin{subfigure}[b]{.18\linewidth}
\centering
  \begin{tabular}{cc}
    ~~~\small $T = 4$\\
     ~~~~~\includegraphics[width=0.9\linewidth]{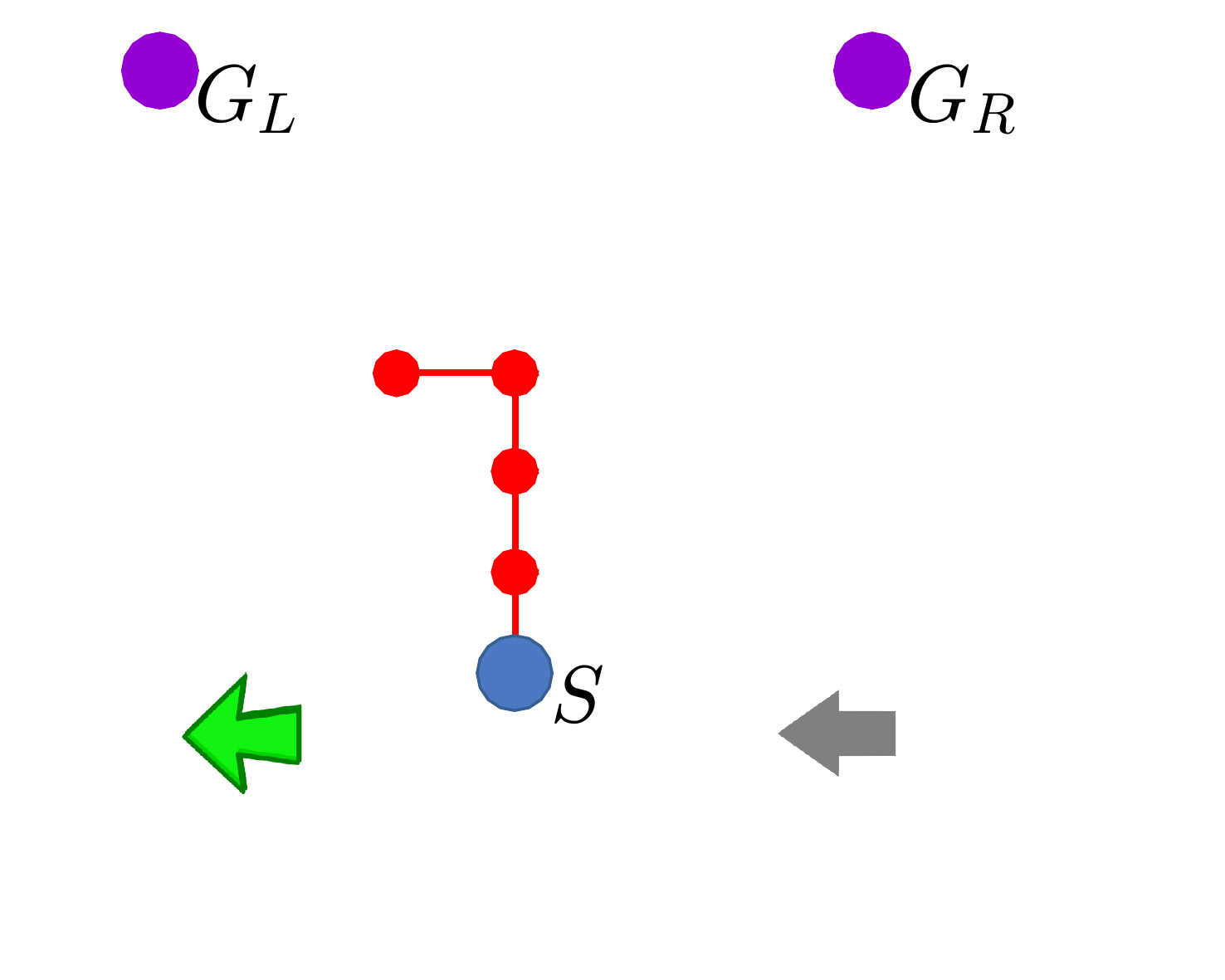}\vspace{-0.3cm}\\
 \includegraphics[width=0.9\linewidth]{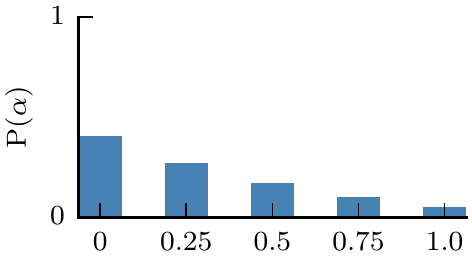}\\
 \includegraphics[width=0.9\linewidth]{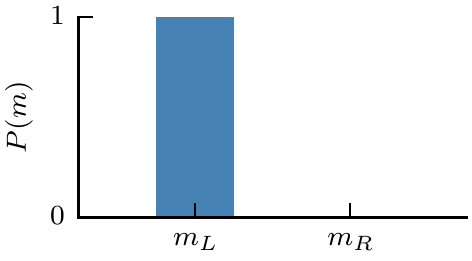}\\
   \end{tabular}
 \label{fig:plotT0}
\end{subfigure}
\begin{subfigure}[b]{.18\linewidth}
\centering
  \begin{tabular}{cc}
    ~~~\small $T = 5$\\
      ~~~~~\includegraphics[width=0.9\linewidth]{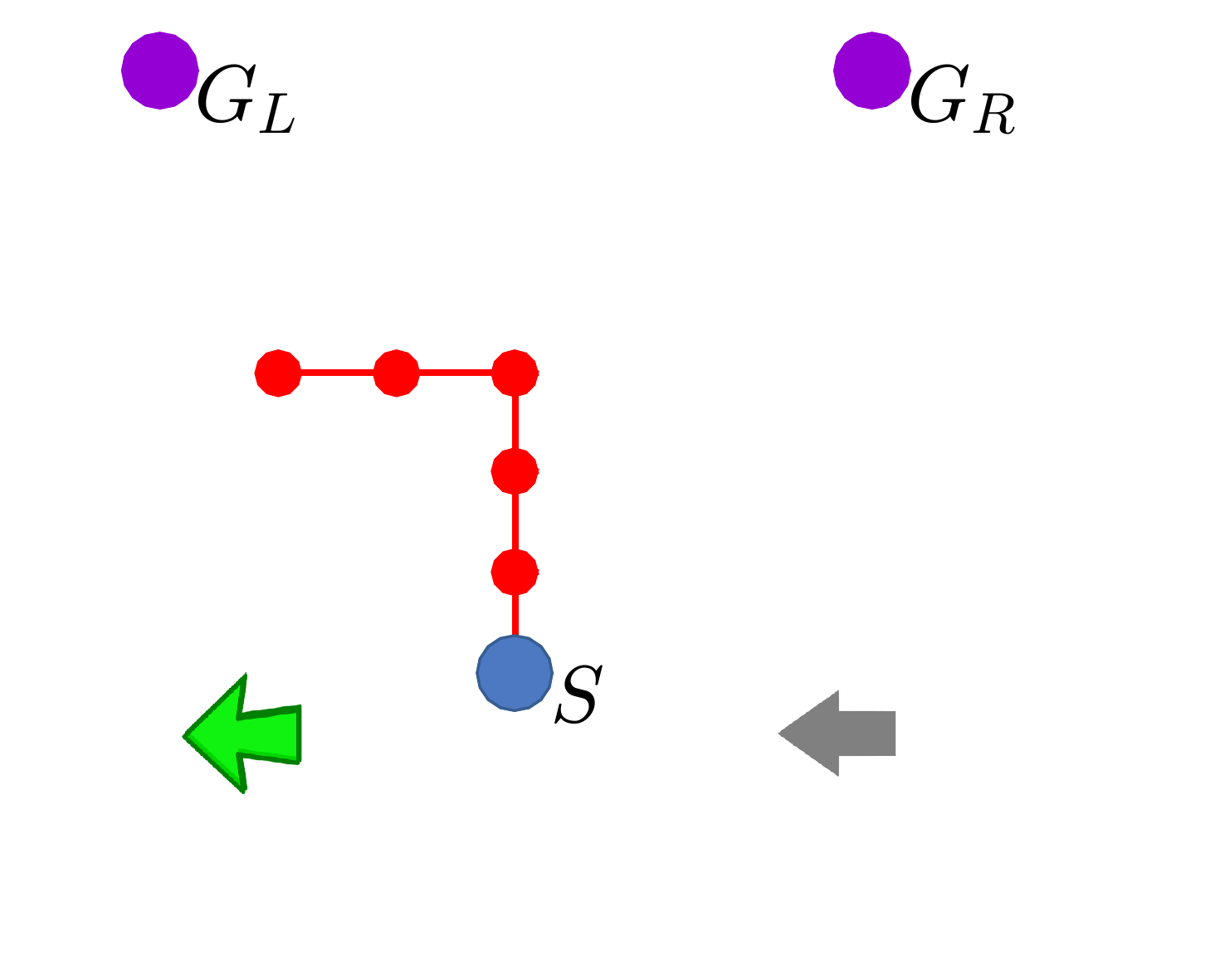}\vspace{-0.3cm}\\
 \includegraphics[width=0.9\linewidth]{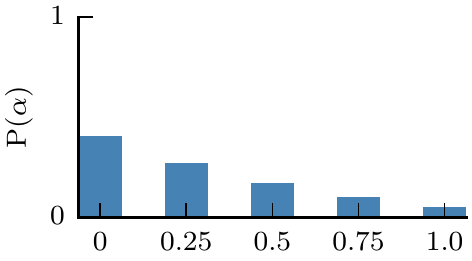}\\
 \includegraphics[width=0.9\linewidth]{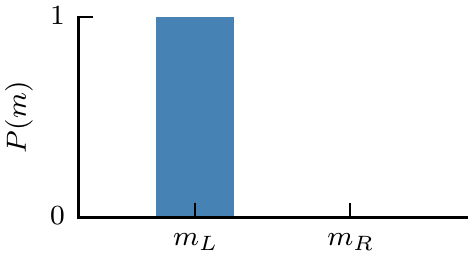}\\

   \end{tabular}
 \label{fig:plotT0}
\end{subfigure}
\end{tabular}
\end{subfigure}

  \vspace{-2.5cm}
\begin{subfigure}[b]{1.0\linewidth}
\begin{tabular}{ccccccc}
\begin{subfigure}[l]{.065\linewidth}
\centering
  \small User 2 \\ $\alpha = 0.75$ $m = m_L$ \\
  \vspace{6cm}
\end{subfigure}
\begin{subfigure}[b]{.18\linewidth}
\centering
  \begin{tabular}{cc}
    ~~~~~\includegraphics[width=0.9\linewidth]{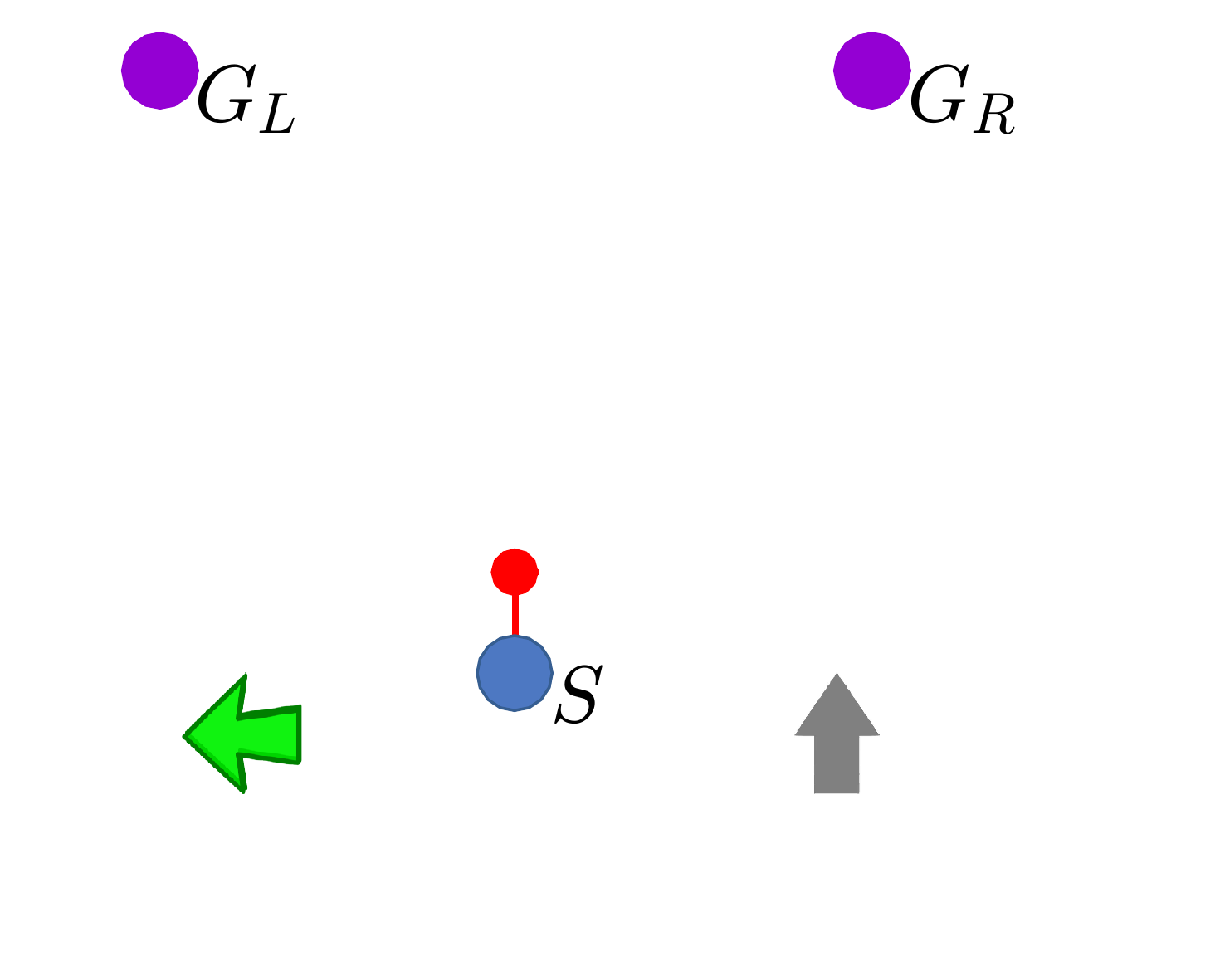}\vspace{-0.3cm}\\
 \includegraphics[width=0.9\linewidth]{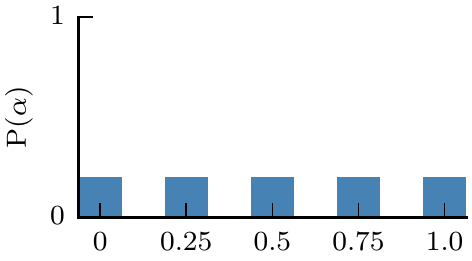}\\
 \includegraphics[width=0.9\linewidth]{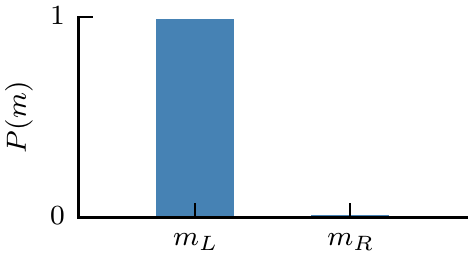}\\
   \end{tabular}
 \label{fig:plotT0}
\end{subfigure}
&
\begin{subfigure}[b]{.18\linewidth}
\centering
  \begin{tabular}{cc}
   ~~~~~\includegraphics[width=0.9\linewidth]{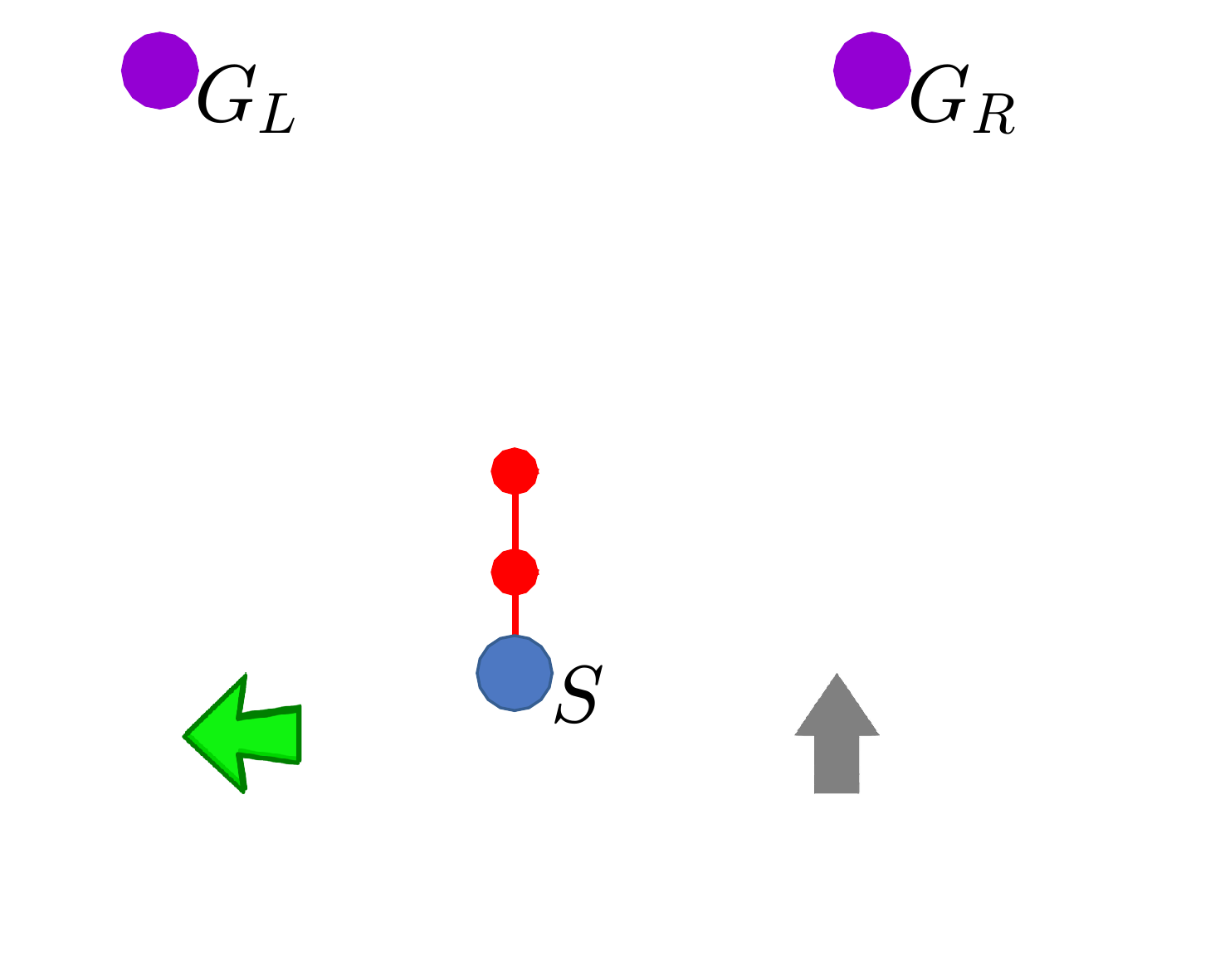}\vspace{-0.3cm}\\
 \includegraphics[width=0.9\linewidth]{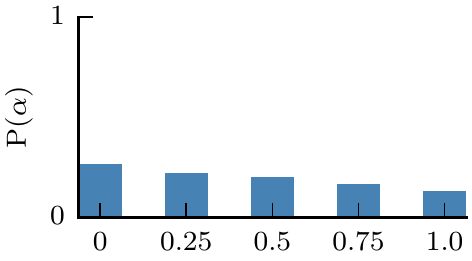}\\
  \includegraphics[width=0.9\linewidth]{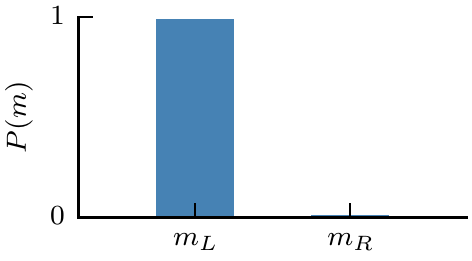}\\
   \end{tabular}
 \label{fig:plotT0}
\end{subfigure}
\begin{subfigure}[b]{.18\linewidth}
\centering
  \begin{tabular}{cc}
      ~~~~~\includegraphics[width=0.9\linewidth]{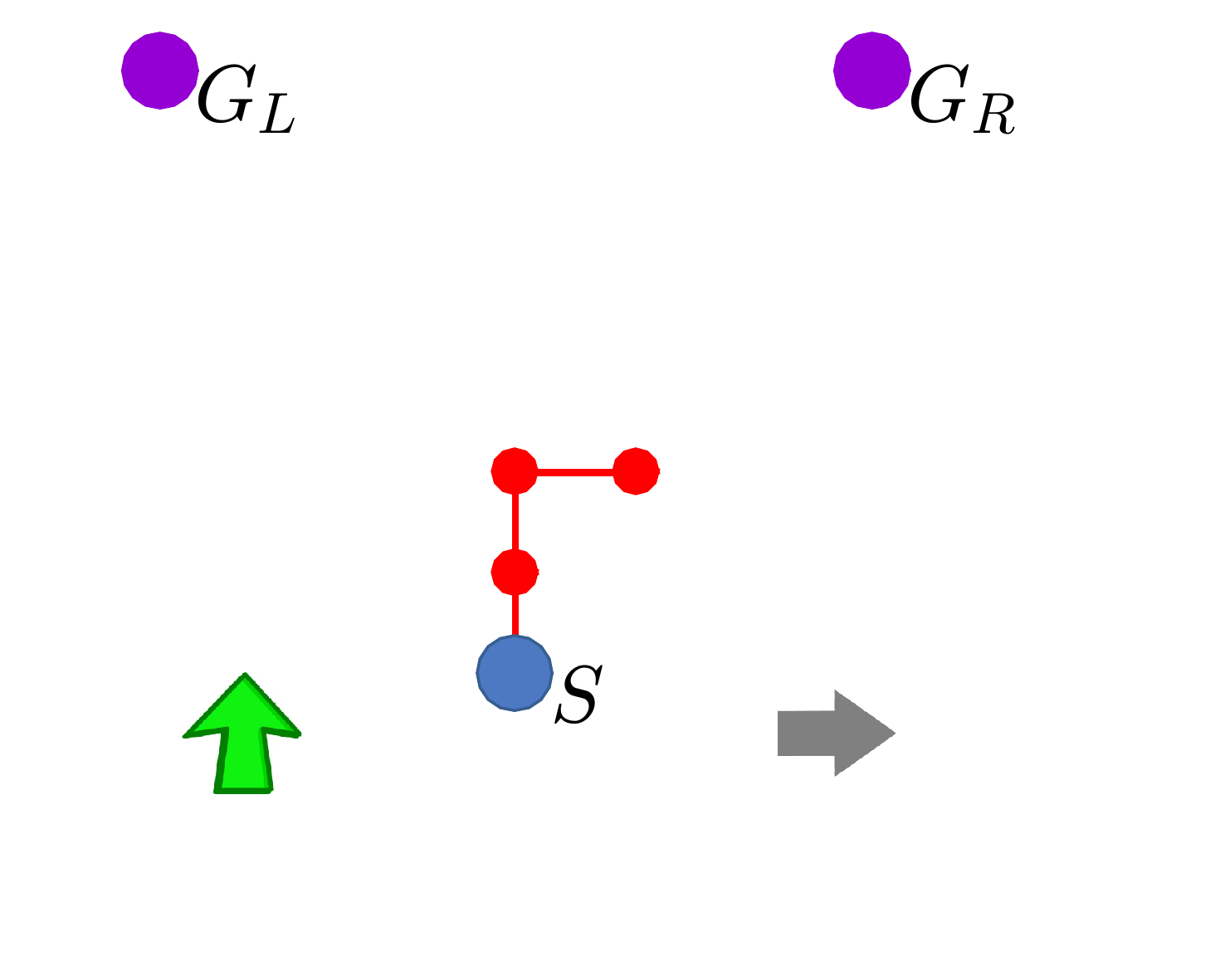}\vspace{-0.3cm}\\
 \includegraphics[width=0.9\linewidth]{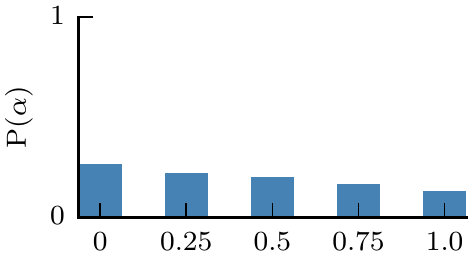}\\
 \includegraphics[width=0.9\linewidth]{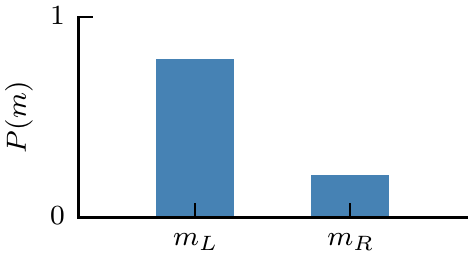}\\
   \end{tabular}
 \label{fig:plotT0}
\end{subfigure}
&
\begin{subfigure}[b]{.18\linewidth}
\centering
  \begin{tabular}{cc}
     ~~~~~\includegraphics[width=0.9\linewidth]{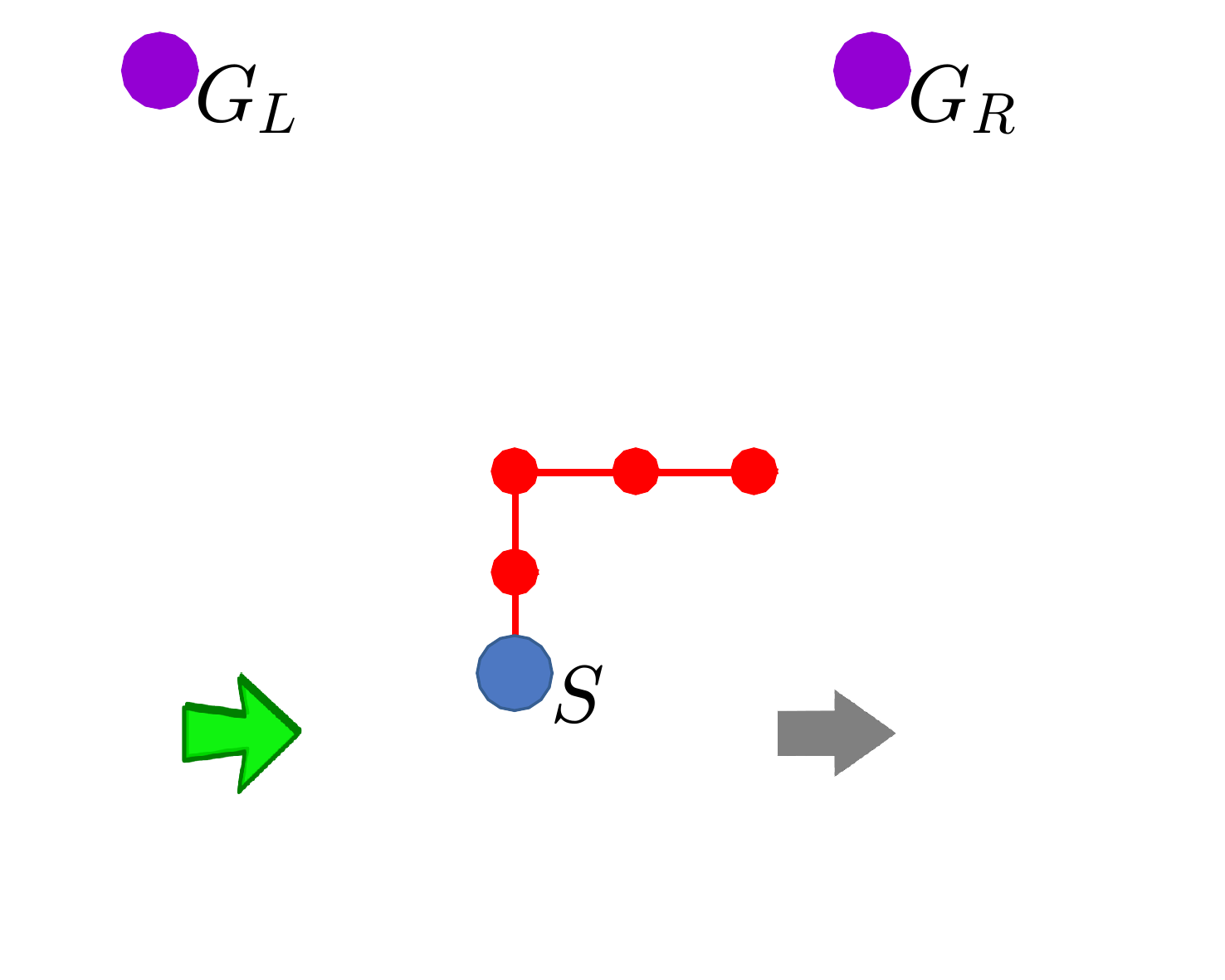}\vspace{-0.3cm}\\
 \includegraphics[width=0.9\linewidth]{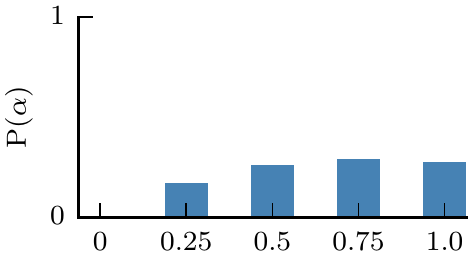}\\
 \includegraphics[width=0.9\linewidth]{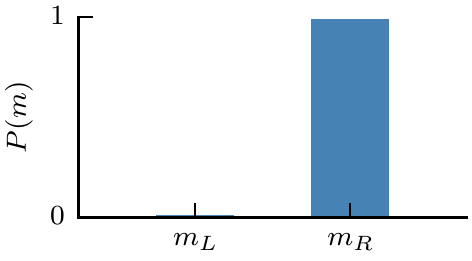}\\
   \end{tabular}
 \label{fig:plotT0}
\end{subfigure}
\begin{subfigure}[b]{.18\linewidth}
\centering
  \begin{tabular}{cc}
      ~~~~~\includegraphics[width=0.9\linewidth]{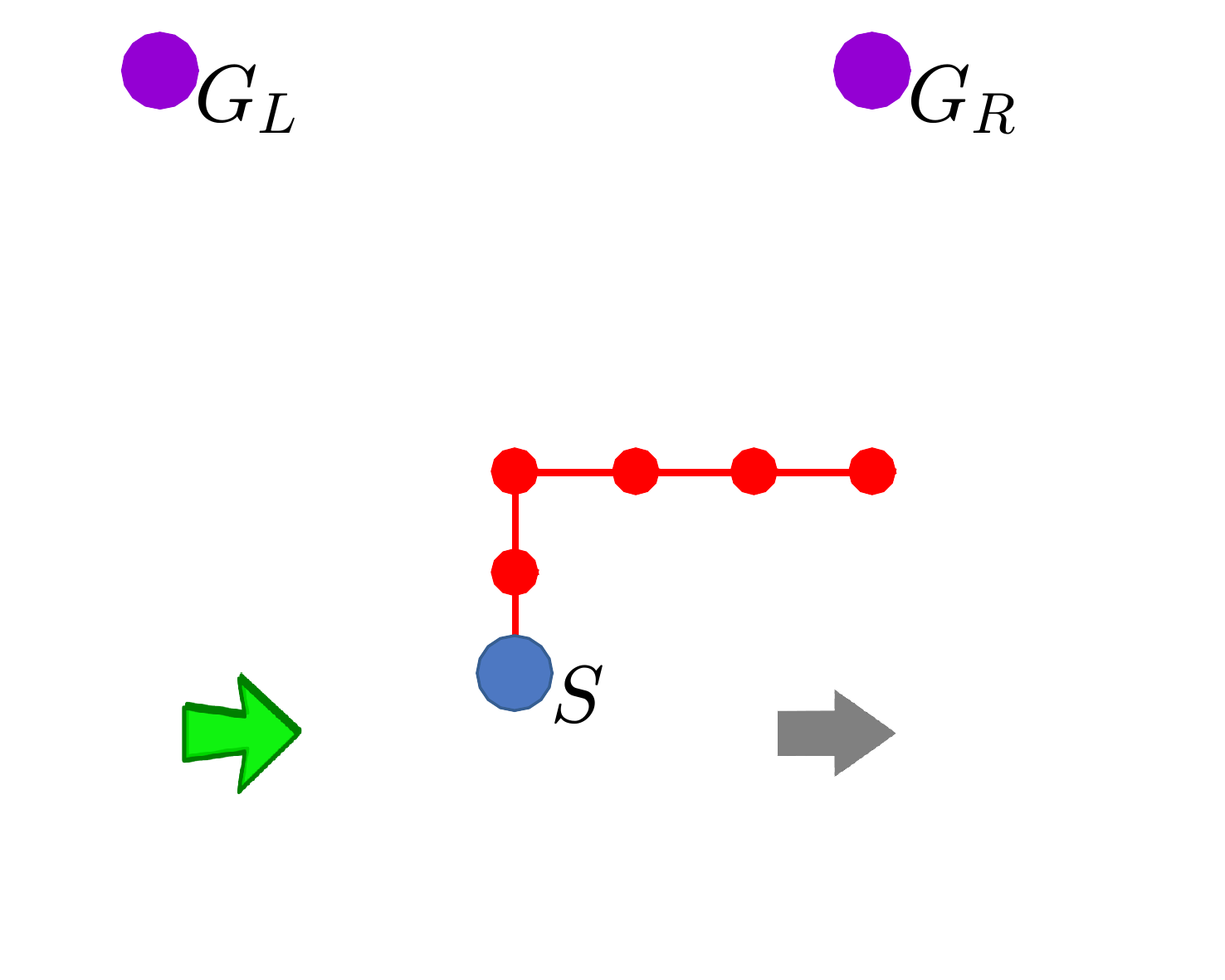}\vspace{-0.3cm}\\
 \includegraphics[width=0.9\linewidth]{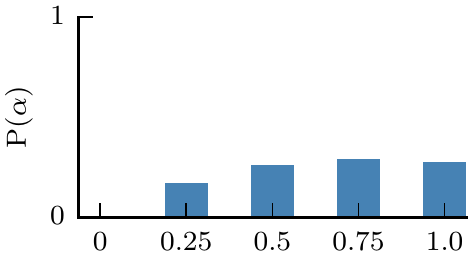}\\
 \includegraphics[width=0.9\linewidth]{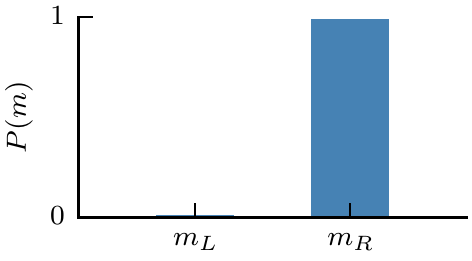}\\

   \end{tabular}
 \label{fig:plotT0}
\end{subfigure}
\end{tabular}
\end{subfigure}

\vspace{-3cm}

\caption{
Sample runs on a shared autonomy scenario with two goals $G_L,G_R$ and two simulated humans of adaptability level $\alpha=0$ and $0.75$. Both users start with modal policy $m_L$ (associated with the left goal). The human and robot actions are \{move-left, move-right, move-forward\}. For both users, the upper row plots the robot trajectory (red dots), the human input (green arrow) and the robot action (gray arrow) over time. The middle row  plots the estimate of $\alpha$ over time, where $\alpha \in \{0,0.25,0.5,0.75,1\}$. Each graph plots $\alpha$ versus the probability of $\alpha$. The lower row plots $m \in \{m_L, m_R\}$ versus the probability of $m$. Columns indicate successive time-steps. User 1 insists on their initial strategy throughout the task and the robot complies, whereas User 2 adapts to the robot and ends up following $m_R$.
}

\label{fig:simulations}
\end{figure*}

\section{Human-Robot Mutual Adaptation Formalism} \label{sec:formalism}

\subsection{MOMDP Formulation} \label{subsec:MOMDP-shared-autonomy}
In Section~\ref{subsec:human-mode-inference}, we showed how the robot estimates the human mode, and how it computes the probability of the human switching to the robot mode based on the human adaptability. In Section~\ref{sec:optimal-disagreement}, we defined a reward function that the robot is maximizing, which captures the trade-off between going to the optimal goal and following the human mode. Both the human adaptability and the human mode are not directly observable. Therefore, the robot needs to estimate them through interaction, while performing the task. This leads us to formulate this problem as a mixed-observability Markov Decision Process (MOMDP)~\cite{ong2010planning}. This formulation allows us to compute an optimal policy for the robot that will maximize the expected reward that the human-robot team will receive, given the robot's estimates of the human adaptability and of the human mode. We define a MOMDP as a tuple $\{X, Y, A_{r}, \mathcal{T}_{x}, \mathcal{T}_{\alpha}, \mathcal{T}_{m_h},R,\Omega, O\}$:
\begin{itemize}

\item $X: X_r \times A_r^{k}$ is the set of observable variables. These are the current robot configuration $x_r$, as well as the history $h_k$. Since $x_r$ transitions deterministically, we only need to register the current robot state and robot actions $a^{t-k+1}_r, ... ,a^{t}_r$.  

\item $Y: \mathcal{A} \times M$ is the set of partially observable variables. These are the human adaptability $\alpha \in A$, and the human mode $m_h \in M$.

\item $A_{r}$ is a finite set of robot actions. We model actions as transitions between discrete robot configurations.

\item $\mathcal{T}_{x}: X \times A_{r} \longrightarrow X$ is a deterministic mapping from a robot configuration $x_r$, history $h_k$ and action $a_r$, to a subsequent configuration $x'_{r}$ and history $h'_k$.

\item $\mathcal{T}_{\alpha}: \mathcal{A} \times A_{r} \longrightarrow \Pi(\mathcal{A})$ is the probability of the human adaptability being $\alpha'$ at the next time step, if the adaptability of the human at time $t$ is $\alpha$ and the robot takes action $a_{r}$. We assume the human adaptability to be fixed throughout the task.

\item $\mathcal{T}_{m_h}: X \times \mathcal{A} \times M\times A_r  \longrightarrow \Pi(M)$ is the probability of the human switching from mode $m_h$ to a new mode $m'_h$, given a history $h_k$, robot state $x_r$, human adaptability $\alpha$ and robot action $a_r$. It is computed using Eq.~\ref{e:m_h-transition}, Sec.~\ref{subsec:human-mode-inference}.

\item $R :  X \times M\times A_r  \longrightarrow \mathbb{R}$ is a reward function that gives an immediate reward for the robot taking action $a_{r}$ given a history $h_k$, human mode $m_h$ and robot state $x_r$. It is defined in Eq.~\ref{e:reward_full}, Sec.~\ref{sec:optimal-disagreement}.

\item $\Omega$ is the set of observations that the robot receives. An observation is a human input $a_h \in A_{h}$ ($\Omega \equiv A_{h}$).  

\item $O : M \times X_r \longrightarrow \Pi(\Omega)$ is the observation function, which gives a probability distribution over human actions for a mode $m_h$ at state $x_r$. This distribution is specified by the stochastic modal policy $m_h \in M$.
\end{itemize}

\subsection{Belief Update}

Based on the above, the belief update for the MOMDP is \cite{ong2010planning}:
\begin{equation}
\label{e:beliefupdate2}
\begin{split}
b'(\alpha', m_h') &= \eta O(m_h',x'_r, a_h) \sum_{\alpha \in \mathcal{A}} \sum_{m_h \in M}\mathcal{T}_{x}(x, a_{r}, x') \\
&\mathcal{T}_{\alpha}(\alpha,a_{r},\alpha')\mathcal{T}_{m_h}(x, \alpha,m_h,a_{r},m'_h)b(\alpha,m_h)\\
\end{split}
\end{equation}

We note that since the MOMDP has two partially observable variables, $\alpha$ and $m_h$, the robot maintains a joint probability distribution over both variables.

\subsection{Robot Policy} \label{e:algorithm}

We solve the MOMDP for a robot policy $\pi^*_r(b)$ that is optimal with respect to the robot's expected total reward. 

The stochastic modal policies may assign multiple actions at a given state. Therefore, even if $m_h \equiv m_r$, $a_r$ may not match the human input $a_h$. Such disagreements are unnecessary when human and robot modes are the same. Therefore, we let the robot actions match the human inputs, if the robot has enough confidence that robot and human modes (computed using Eq.~\ref{e:m_r_simple},~\ref{e:beliefupdate2a}) are identical in the current time-step. Otherwise, the robot executes the action specified by the MOMDP optimal policy. We leave for future work adding a penalty for disagreement between actions, which we hypothesize it would result in similar behavior. 

\begin{figure}[t!]
\centering
  \includegraphics[width=0.55\linewidth]{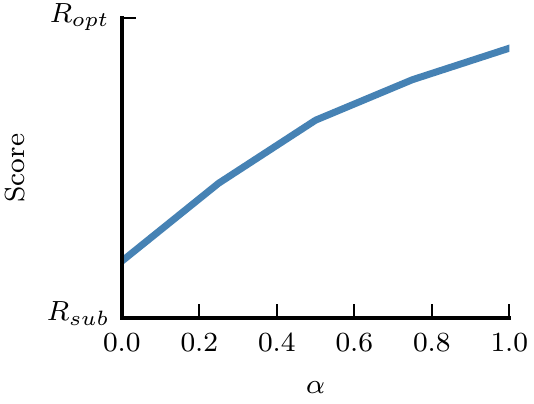}
 \caption{Mean performance for simulated users of different adaptability $\alpha$.}
 \label{fig:performance-over-alpha}
\end{figure}

\subsection{Simulations}
Fig.~\ref{fig:simulations} shows the robot behavior for two simulated users, one with low adaptability (User 1, $\alpha = 0.0$), and one with high adaptability (User 2, $\alpha = 0.75$) for a shared autonomy scenario with two goals, $G_L$ and $G_R$, corresponding to modal policies $m_L$ and $m_R$ respectively. Both users start with modal policy $m_L$ (left goal). The robot uses the human input to estimate both $m_h$ and $\alpha$. We set a bounded-memory of $k=1$ time-step. If human and robot disagree and the human insists on their modal policy, then the MOMDP belief is updated so that smaller values of adaptability $\alpha$ have higher probability (lower adaptability). It the human aligns its inputs to the robot mode, larger values become more likely. If the robot infers the human to be  adaptable, it moves towards the optimal goal. Otherwise, it complies with the human, thus retaining their trust.

Fig.~\ref{fig:performance-over-alpha} shows the team-performance over $\alpha$, averaged over 1000 runs with simulated users. We evaluate performance by the reward of the goal achieved, where $R_{opt}$ is the reward for the optimal and $R_{sub}$ for the sub-optimal goal. We see that the more adaptable the user, the more often the robot will reach the optimal goal. Additionally, we observe that for $\alpha = 0.0$, the performance is higher than $R_{sub}$. This is because the simulated user may choose to move forward in the first time-steps; when the robot infers that they are stubborn, it is already close to the optimal goal and continues moving to that goal.

\section{Human Subject Experiment} \label{sec:experiments}
We conduct a human subject experiment ($n=51$) in a shared autonomy setting. We are interested in showing that the human-robot mutual adaptation formalism can improve the performance of human-robot teams, while retaining high levels of perceived collaboration and trust in the robot in the shared autonomy domain. 

On one extreme, we ``fix'' the robot policy, so that the robot always moves towards the optimal goal, ignoring human adaptability. We hypothesize that this will have a negative effect on human trust and perceived robot performance as a teammate. On the other extreme, we have the robot assist the human in achieving their desired goal.
 

We show that the proposed formalism achieves a trade-off between the two: when the human is non-adaptable, the robot follows the human preference. Otherwise, the robot insists on the optimal way of completing the task, leading to significantly better policies, compared to following the human preference, while achieving a high level of trust.

\subsection{Independent Variables}
\noindent\textbf{No-adaptation session.} The robot executes a fixed policy, always acting towards the optimal goal.

\noindent\textbf{Mutual-adaptation session.} The robot executes the MOMDP policy of Sec. \ref{e:algorithm}. 

\noindent\textbf{One-way adaptation session.} The robot estimates a distribution over user goals, and adapts to the user following their preference, assisting them for that distribution~\cite{Javdani-RSS-15}. We compute the robot policy in that condition by fixing the adaptability value to 0 in our model and assigning equal reward to both goals. 
\\
\subsection{Hypotheses}
\noindent\textbf{H1} \textit{The performance of teams in the No-adaptation condition will be better than of teams in the Mutual-adaptation condition, which will in turn be better than of teams in the One-way adaptation condition.} We expected teams in the No-adaptation condition to outperform the teams in the other conditions, since the robot will always go to the optimal goal. In the Mutual-adaptation condition, we expected a significant number of users to adapt to the robot and switch their strategy towards the optimal goal. Therefore, we posited that this would result in an overall higher reward, compared to the reward resulting from the robot following the participants' preference throughout the task (One-way adaptation). 

\noindent\textbf{H2} \textit{Participants that work with the robot in the One-way adaptation condition will rate higher their trust in the robot, as well as their perceived collaboration with the robot, compared to working with the robot in the Mutual-adaptation condition,. Additionally, participants in the Mutual-adaptation condition will give higher ratings, compared to working with the robot in the No-adaptation condition.} 
We expected users to trust the robot more in the One-way adaptation condition than in the other conditions, since in that condition the robot will always follow their preference. In the Mutual-adaptation condition, we expected users to trust the robot more and perceive it as a better teammate, compared with the robot that executed a fixed strategy ignoring users' adaptability (No-adaptation). Previous work in collaborative tasks has shown a significant improvement in human trust, when the robot had the ability to adapt to the human parter~\cite{nikolaidis2016formalizing, shah2011improved, lasota2015analyzing}



\subsection{Experiment Setting: A Table Clearing Task}
Participants were asked to clear a table off two bottles placed symmetrically, by providing inputs to a robotic arm through a joystick interface (Fig.~\ref{fig:screenshot}). They controlled the robot in Cartesian space by moving it in three different directions: left, forward and right. We first instructed them in the task, and asked them to do two training sessions, where they practiced controlling the robot with the joystick.  We then asked them to choose which of the two bottles they would like the robot to grab first, and we set the robot policy, so that the other bottle was the optimal goal. This emulates a scenario where, for instance, the robot would be unable to grasp one bottle without dropping the other, or where one bottle would be heavier than the other and should be placed in the bin first. In the one-way and mutual adaptation conditions, we told them that ``the robot has a mind of its own, and it may choose not to follow your inputs.'' Participants then did the task three times in all conditions, and then answered a post-experimental questionnaire that used a five-point Likert scale to assess their responses to working with the robot. Additionally, in a video-taped interview at the end of the task, we asked participants that had changed strategy during the task to justify their action.

\begin{figure*}[t!]
\centering
  \includegraphics[width=0.85\linewidth]{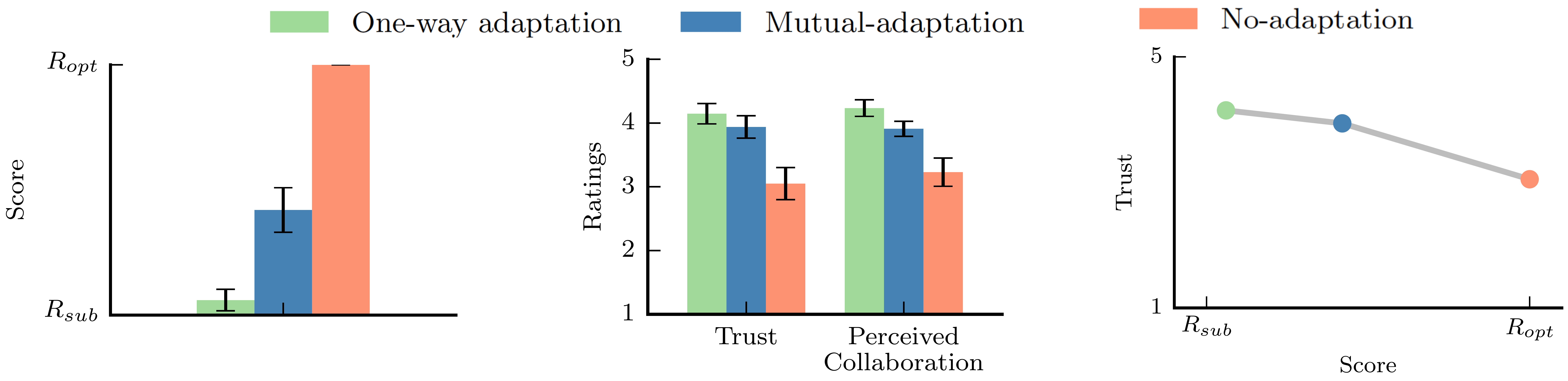}
 \caption{Findings for objective and subjective measures.}
 \label{fig:trust}
\end{figure*}

\subsection{Subject Allocation}
 We recruited 51 participants from the local community, and chose a between-subjects design in order to not bias the users with policies from previous conditions. 

 \subsection{MOMDP Model} \label{subsec:MOMDPModel}
The size of the observable state-space $X$ was 52 states. We empirically found that a history length of $k=1$ in BAM was sufficient for this task, since most of the subjects that changed their preference did so reacting to the previous robot action. The human and robot actions were \{move-left, move-right, move-forward\}. We specified two stochastic modal policies $\{m_L, m_R\}$, one for each goal. We additionally assumed a discrete set of values of the adaptability $\alpha$ : $\{0.0,0.25,0.5,0.75,1.0\}$. Therefore, the total size of the MOMDP state-space was $5 \times 2 \times 52 = 520$ states. We selected the reward so that $R_{opt} = 11$ for the optimal goal, $R_{sub} = 10$ for the suboptimal goal, and $C = -0.32$ for the cost of mode disagreement (Eq.~\ref{e:disagreement_cost}). We computed the robot policy using the SARSOP solver~\cite{kurniawati2008sarsop}, a point-based approximation algorithm which, combined with the MOMDP formulation, can scale up to hundreds of thousands of states~\cite{bandyopadhyay2013intention}.

\section{Analysis}
\subsection{Objective Measures}
We consider hypothesis \noindent\textbf{H1}, that the performance of teams in the No-adaptation condition will be better than of teams in the Mutual-adaptation condition, which in turn will be better than of teams in the One-way adaptation condition. 

Nine participants out of 16 ($56\%$) in the Mutual-adaptation condition guided the robot towards the optimal goal, which was different than their initial preference, during the final trial of the task, while 12 out of 16 ($75\%$) did so at one or more of the three trials. From the participants that changed their preference, only one stated that they did so for reasons irrelevant to the robot policy. On the other hand, only two participants out of 17 in the One-way adaptation condition changed goals during the task, while 15 out of 17 guided the robot towards their preferred, suboptimal goal in all trials. This indicates that the adaptation observed in the Mutual-adaptation condition was caused by the robot behavior.


We evaluate team performance by computing the mean reward over the three trials, with the reward for each trial being $R_{opt}$ if the robot reached the optimal goal and $R_{sub}$ if the robot reached the suboptimal goal (Fig.~\ref{fig:trust}-left). As expected, a Kruskal-Wallis H test showed that there was a statistically significant difference in performance among the different conditions ($\chi^2(2) = 39.84, p < 0.001$). Pairwise two-tailed Mann-Whitney-Wilcoxon tests with Bonferroni corrections showed the difference to be statistically significant between the No-adaptation and Mutual-adaptation ($U = 28.5, p < 0.001$), and Mutual-adaptation and One-way adaptation  ($U = 49.5, p = 0.001$) conditions. This supports our hypothesis. 


\subsection{Subjective Measures}
Recall hypothesis \textbf{H2}, that participants in the Mutual-adaptation condition would rate their trust and perceived collaboration with the robot higher than in the No-adaptation condition, but lower than in the One-way adaptation condition. Table I shows the two subjective scales that we used. The \textit{trust} scales were used as-is from~\cite{hoffman2013evaluating}. We additionally chose a set of questions related to participants' \textit{perceived collaboration} with the robot.

Both scales had good consistency. Scale items were combined into a score. Fig.~\ref{fig:trust}-center shows that both participants' trust ($M=3.94, SE=0.18$) and perceived collaboration ($M=3.91, SE=0.12$) were high in the Mutual-adaptation condition. One-way ANOVAs showed a statistically significant difference between the three conditions in both trust  ($F(2,48)=8.370, p = 0.001$) and perceived collaboration ($F(2,48)=9.552, p < 0.001$). Tukey post-hoc tests revealed that participants of the Mutual-adaptation condition trusted the robot more, compared to participants that worked with the robot in the No-adaptation condition ($p = 0.010$). Additionally, they rated higher their perceived collaboration with the robot ($p = 0.017$). However, there was no significant difference in either measure between participants in the One-way adaptation and Mutual-adaptation conditions. We attribute these results to the fact that the MOMDP formulation allowed the robot to reason over its estimate of the adaptability of its teammate; if the teammate insisted towards the suboptimal goal, the robot responded to the input commands and followed the user's preference. If the participant changed their inputs based on the robot actions, the robot guided them towards the optimal goal, while retaining a high level of trust. By contrast, the robot in the No-adaptation condition always moved towards the optimal goal ignoring participants' inputs, which in turn had a negative effect on subjective measures.




\section{Discussion}
In this work, we proposed a human-robot mutual adaptation formalism in a shared autonomy setting. In a human subject experiment, we compared the policy computed with our formalism, with an assistance policy, where the robot helped participants to achieve their intended goal, and with a fixed policy where the robot always went towards the optimal goal. 

As Fig.~\ref{fig:trust} illustrates, participants in the one-way adaptation condition had the worst performance, since they guided the robot towards a suboptimal goal. The fixed policy achieved maximum performance, as expected. However, this came to the detriment of human trust in the robot. On the other hand, the assistance policy in the One-way adaptation condition resulted in the highest trust ratings --- albeit not significantly higher than the ratings in the Mutual-adaptation condition --- since the robot always followed the user preference and there was no goal disagreement between human and robot. Mutual-adaptation balanced the trade-off between optimizing performance and retaining trust: users in that condition trusted the robot more than in the No-adaptation condition, and performed better than in the One-way adaptation condition.

Fig.~\ref{fig:trust}-right shows the three conditions with respect to trust and performance scores. We can make the MOMDP policy identical to either of the two policies in the end-points, by changing the MOMDP model parameters. If we fix in the model the human adaptability to 0 and assign equal costs for both goals, the robot would assist the user in their goal (One-way adaptation). If we fix adaptability to 1 in the model (or we remove the penalty for mode disagreement), the robot will always go to the optimal goal (fixed policy).

\begin{table}[t!]
\centering
\textsc{Table I: Subjective Measures}
\begin{tabular}{p{0.9\columnwidth}}
\\ \toprule\hline
\textbf{Trust} $\alpha=.85$\\
\emph{1.I trusted the robot to do the right thing at the right time.}\\
\emph{2.The robot was trustworthy.}\\
\hline
\textbf{Perceived Collaboration} $\alpha=.81$\\
\emph{1.I was satisfied with ADA and my performance.}\\
\emph{2.The robot and I worked towards mutually agreed upon goals.}\\
\emph{3.The robot and I collaborated well together}\\
\emph{4.The robot's actions were reasonable.} \\ 
\emph{5.The robot was responsive to me.} \\ 

\hline
\bottomrule
\end{tabular}
\end{table}

The presented table-clearing task can be generalized without significant modifications to tasks with a large number of goals, human inputs and robot actions, such as picking good grasps in manipulation tasks (Fig.~\ref{fig:grasp-strategy}): The state-space size increases linearly with $(1/dt)$, where $dt$ a discrete time-step, and with the number of modal policies. On the other hand, the number of observable states is polynomial to the number of robot actions ($O(A_r^k))$, since each state includes history $h_k$: For tasks with large $|A_r|$ and memory length $k$, we could approximate $h_k$ using feature-based representations.


Overall, we are excited to have brought about a better understanding of the relationships between adaptability, performance and trust in a shared autonomy setting. We are very interested in exploring applications of these ideas beyond assistive robotic arms, to powered wheelchairs, remote manipulators, and generally to settings where human inputs are combined with robot autonomy.

\section*{Acknowledgments}
{ \small 
We thank Michael Koval, Shervin Javdani and Henny Admoni for the very helpful discussion and advice.}
\section*{Funding}
{\small
 This work was funded by the DARPA SIMPLEX program through ARO contract number 67904LSDRP, National Institute of Health R01 (\#R01EB019335), National Science Foundation CPS (\#1544797), and the Office of Naval Research. We also acknowledge the Onassis Foundation as a sponsor.
}

\balance
\bibliographystyle{IEEEtran}
\bibliography{mybib2short}  

\begin{thebibliography}{10}
\providecommand{\url}[1]{#1}
\csname url@samestyle\endcsname
\providecommand{\newblock}{\relax}
\providecommand{\bibinfo}[2]{#2}
\providecommand{\BIBentrySTDinterwordspacing}{\spaceskip=0pt\relax}
\providecommand{\BIBentryALTinterwordstretchfactor}{4}
\providecommand{\BIBentryALTinterwordspacing}{\spaceskip=\fontdimen2\font plus
\BIBentryALTinterwordstretchfactor\fontdimen3\font minus
  \fontdimen4\font\relax}
\providecommand{\BIBforeignlanguage}[2]{{%
\expandafter\ifx\csname l@#1\endcsname\relax
\typeout{** WARNING: IEEEtran.bst: No hyphenation pattern has been}%
\typeout{** loaded for the language `#1'. Using the pattern for}%
\typeout{** the default language instead.}%
\else
\language=\csname l@#1\endcsname
\fi
#2}}
\providecommand{\BIBdecl}{\relax}
\BIBdecl

\bibitem{hillman2002weston}
M.~Hillman, K.~Hagan, S.~Hagan, J.~Jepson, and R.~Orpwood, ``The weston
  wheelchair mounted assistive robot \- the design story,'' \emph{Robotica},
  vol.~20, no.~02, pp. 125--132, 2002.

\bibitem{prior1990electric}
S.~D. Prior, ``An electric wheelchair mounted robotic arm \- a survey of
  potential users,'' \emph{Journal of medical engineering \& technology},
  vol.~14, no.~4, pp. 143--154, 1990.

\bibitem{sijs2007combined}
J.~Sijs, F.~Liefhebber, and G.~W.~R. Romer, ``Combined position \& force
  control for a robotic manipulator,'' in \emph{2007 IEEE 10th International
  Conference on Rehabilitation Robotics}.\hskip 1em plus 0.5em minus
  0.4em\relax IEEE, 2007, pp. 106--111.

\bibitem{kofman2005teleoperation}
J.~Kofman, X.~Wu, T.~J. Luu, and S.~Verma, ``Teleoperation of a robot
  manipulator using a vision-based human-robot interface,'' \emph{Industrial
  Electronics, IEEE Transactions on}, vol.~52, no.~5, pp. 1206--1219, 2005.

\bibitem{dragan2013policy}
A.~D. Dragan and S.~S. Srinivasa, ``A policy-blending formalism for shared
  control,'' \emph{The International Journal of Robotics Research}, vol.~32,
  no.~7, pp. 790--805, 2013.

\bibitem{yu2005telemanipulation}
W.~Yu, R.~Alqasemi, R.~Dubey, and N.~Pernalete, ``Telemanipulation assistance
  based on motion intention recognition,'' in \emph{Robotics and Automation,
  2005. ICRA 2005. Proceedings of the 2005 IEEE International Conference
  on}.\hskip 1em plus 0.5em minus 0.4em\relax IEEE, 2005, pp. 1121--1126.

\bibitem{trautman2015assistive}
P.~Trautman, ``Assistive planning in complex, dynamic environments: a
  probabilistic approach,'' in \emph{Systems, Man, and Cybernetics (SMC), 2015
  IEEE International Conference on}.\hskip 1em plus 0.5em minus 0.4em\relax
  IEEE, 2015, pp. 3072--3078.

\bibitem{gopinath2017human}
D.~Gopinath, S.~Jain, and B.~D. Argall, ``Human-in-the-loop optimization of
  shared autonomy in assistive robotics,'' \emph{IEEE Robotics and Automation
  Letters}, vol.~2, no.~1, pp. 247--254, 2017.

\bibitem{Javdani-RSS-15}
S.~Javdani, S.~Srinivasa, and J.~A.~D. Bagnell, ``Shared autonomy via hindsight
  optimization,'' in \emph{Proceedings of Robotics: Science and Systems}, Rome,
  Italy, July 2015.

\bibitem{hancock2011meta}
P.~A. Hancock, D.~R. Billings, K.~E. Schaefer, J.~Y. Chen, E.~J. De~Visser, and
  R.~Parasuraman, ``A meta-analysis of factors affecting trust in human-robot
  interaction,'' \emph{Human Factors}, 2011.

\bibitem{salem2015would}
M.~Salem, G.~Lakatos, F.~Amirabdollahian, and K.~Dautenhahn, ``Would you trust
  a (faulty) robot?: Effects of error, task type and personality on human-robot
  cooperation and trust,'' in \emph{HRI}, 2015.

\bibitem{lee2013computationally}
J.~J. Lee, W.~B. Knox, J.~B. Wormwood, C.~Breazeal, and D.~DeSteno,
  ``Computationally modeling interpersonal trust,'' \emph{Front. Psychol.},
  2013.

\bibitem{xu2009woz}
Y.~Xu, K.~Ueda, T.~Komatsu, T.~Okadome, T.~Hattori, Y.~Sumi, and T.~Nishida,
  ``Woz experiments for understanding mutual adaptation,'' \emph{Ai \&
  Society}, vol.~23, no.~2, pp. 201--212, 2009.

\bibitem{komatsu2005experiments}
T.~Komatsu, A.~Ustunomiya, K.~Suzuki, K.~Ueda, K.~Hiraki, and N.~Oka,
  ``Experiments toward a mutual adaptive speech interface that adopts the
  cognitive features humans use for communication and induces and exploits
  users' adaptations,'' \emph{International Journal of Human-Computer
  Interaction}, vol.~18, no.~3, pp. 243--268, 2005.

\bibitem{nikolaidis2016formalizing}
S.~Nikolaidis, A.~Kuznetsov, D.~Hsu, and S.~Srinivasa, ``Formalizing
  human-robot mutual adaptation: A bounded memory model,'' in \emph{The
  Eleventh ACM/IEEE International Conference on Human Robot Interation}.\hskip
  1em plus 0.5em minus 0.4em\relax IEEE Press, 2016, pp. 75--82.

\bibitem{ong2010planning}
S.~C. Ong, S.~W. Png, D.~Hsu, and W.~S. Lee, ``Planning under uncertainty for
  robotic tasks with mixed observability,'' \emph{IJRR}, 2010.

\bibitem{bandyopadhyay2013intention}
T.~Bandyopadhyay, K.~S. Won, E.~Frazzoli, D.~Hsu, W.~S. Lee, and D.~Rus,
  ``Intention-aware motion planning,'' in \emph{WAFR}.\hskip 1em plus 0.5em
  minus 0.4em\relax Springer, 2013.

\bibitem{simon1979rational}
H.~A. Simon, ``Rational decision making in business organizations,'' \emph{The
  American economic review}, pp. 493--513, 1979.

\bibitem{powers2005learning}
R.~Powers and Y.~Shoham, ``Learning against opponents with bounded memory.'' in
  \emph{IJCAI}, 2005.

\bibitem{monte2014learning}
D.~Monte, ``Learning with bounded memory in games,'' \emph{GEB}, 2014.

\bibitem{aumann1989cooperation}
R.~J. Aumann and S.~Sorin, ``Cooperation and bounded recall,'' \emph{GEB},
  1989.

\bibitem{ziebart2009planning}
B.~D. Ziebart, N.~Ratliff, G.~Gallagher, C.~Mertz, K.~Peterson, J.~A. Bagnell,
  M.~Hebert, A.~K. Dey, and S.~Srinivasa, ``Planning-based prediction for
  pedestrians,'' in \emph{IROS}, 2009.

\bibitem{ziebart2008maximum}
B.~D. Ziebart, A.~L. Maas, J.~A. Bagnell, and A.~K. Dey, ``Maximum entropy
  inverse reinforcement learning.'' in \emph{AAAI}, 2008, pp. 1433--1438.

\bibitem{shah2011improved}
J.~Shah, J.~Wiken, B.~Williams, and C.~Breazeal, ``Improved human-robot team
  performance using chaski, a human-inspired plan execution system,'' in
  \emph{HRI}, 2011.

\bibitem{lasota2015analyzing}
P.~A. Lasota and J.~A. Shah, ``Analyzing the effects of human-aware motion
  planning on close-proximity human--robot collaboration,'' \emph{Hum.
  Factors}, 2015.

\bibitem{kurniawati2008sarsop}
H.~Kurniawati, D.~Hsu, and W.~S. Lee, ``Sarsop: Efficient point-based pomdp
  planning by approximating optimally reachable belief spaces.'' in \emph{RSS},
  2008.

\bibitem{hoffman2013evaluating}
G.~Hoffman, ``Evaluating fluency in human-robot collaboration,'' in \emph{HRI
  workshop on human robot collaboration}, vol. 381, 2013, pp. 1--8.

\end{thebibliography}
\end{document}